\documentclass[10pt,twocolumn,letterpaper]{article}

\usepackage{iccv}
\usepackage{times}
\usepackage{epsfig}
\usepackage{graphicx}
\usepackage{amsmath}
\usepackage{amssymb}
\usepackage{multirow}
\usepackage{booktabs}
\usepackage{color}
\usepackage{soul}
\usepackage{amssymb}
\usepackage{algorithm}
\usepackage{algpseudocode}
\usepackage{makecell}
\usepackage{placeins}

\usepackage[accsupp]{axessibility}

\definecolor{darkgreen}{RGB}{30,150,30}
\definecolor{darkblue}{RGB}{0,0,127}
\definecolor{darkyellow}{RGB}{171,133,0}
\definecolor{darkred}{RGB}{180,20,20}
\definecolor{darkmagenta}{RGB}{127,0,127}
\definecolor{darkcyan}{RGB}{0,127,127}

\usepackage[breaklinks=true,colorlinks=true,pagebackref=true,bookmarks=false]{hyperref}

\iccvfinalcopy 

\def\iccvPaperID{3399} 

\newcommand{\expectation}{\mathop{\mathbb{E}}}
\DeclareMathOperator*{\argmin}{arg\,min}
\usepackage{bbm}
\usepackage[table,dvipsnames,svgnames,x11names]{xcolor}
\usepackage{lipsum}

\definecolor{ourdarkblue}{rgb}{0.0, 0.53, 0.74}
\definecolor{coolblack}{rgb}{0.0, 0.23, 0.64}
\newcommand{\jnkc}[1]{\textcolor{coolblack}{#1}} 

\algnewcommand\algorithmicforeach{\textbf{for each}}
\algdef{S}[FOR]{ForEach}[1]{\algorithmicforeach\ #1\ \algorithmicdo}

\usepackage{array}
\newcolumntype{P}[1]{>{\centering\arraybackslash}p{#1}}

\algdef{S}[FOR]{ForEach}[1]{\algorithmicforeach\ #1\ \algorithmicdo}

\usepackage{xcolor}
\usepackage{algpseudocode}
\usepackage{algorithm}
\algnewcommand{\algorithmicgoto}{\textbf{go to}}%
\algnewcommand{\Goto}[1]{\algorithmicgoto~\ref{#1}}%

\begin{document}

\title{{Generalize then Adapt: Source-Free Domain Adaptive Semantic Segmentation}}

\author{Jogendra Nath Kundu$^1$\thanks{Equal contribution.} \enspace Akshay Kulkarni$^1$\footnotemark[1] \enspace Amit Singh$^1$ \enspace Varun Jampani$^2$ \enspace R. Venkatesh Babu$^1$ \\
$^1$Indian Institute of Science, Bangalore  \qquad $^2$Google Research\\
}

\maketitle


\begin{abstract}
    Unsupervised domain adaptation (DA) has gained substantial interest in semantic segmentation. However, almost all prior arts assume concurrent access to both labeled source and unlabeled target, making them unsuitable for scenarios demanding source-free adaptation. In this work\footnote{Project page: \url{https://sites.google.com/view/sfdaseg}}, we enable source-free DA by partitioning the task into two: a) source-only domain generalization and b) source-free target adaptation. 
    Towards the former, we provide theoretical insights to develop a multi-head framework trained with a virtually extended multi-source dataset, aiming to balance generalization and specificity.
    Towards the latter, we utilize the multi-head framework to extract reliable target pseudo-labels for self-training. Additionally, we introduce a novel conditional prior-enforcing auto-encoder that discourages spatial irregularities, thereby enhancing the pseudo-label quality. Experiments on the standard GTA5$\to$Cityscapes and SYNTHIA$\to$Cityscapes benchmarks show our superiority even against the non-source-free prior-arts. 
    Further, we show our compatibility with online adaptation enabling deployment in a sequentially changing environment.
\end{abstract}

\section{Introduction}
Almost all supervised learning systems assume that the training and testing data follow the same input distribution. However, this assumption is impractical as target scenarios often exhibit a distribution shift. For example, self-driving cars often fail to generalize when deployed in conditions different from training, such as cross-city \cite{chen2017no} or cross-weather \cite{SDHV18} deployment.
This is because the model fails to apprehend the generic, causal factors of variations and instead, holds on to domain-specific spurious correlations \cite{ilse2021selecting}.
Over-reliance on training data from a particular distribution can cause the model to fail even for mild domain-shifts like changes in illumination, texture, background, \etc

\begin{figure}
    \centering
    \vspace{-2mm}
    \includegraphics[width=\columnwidth]{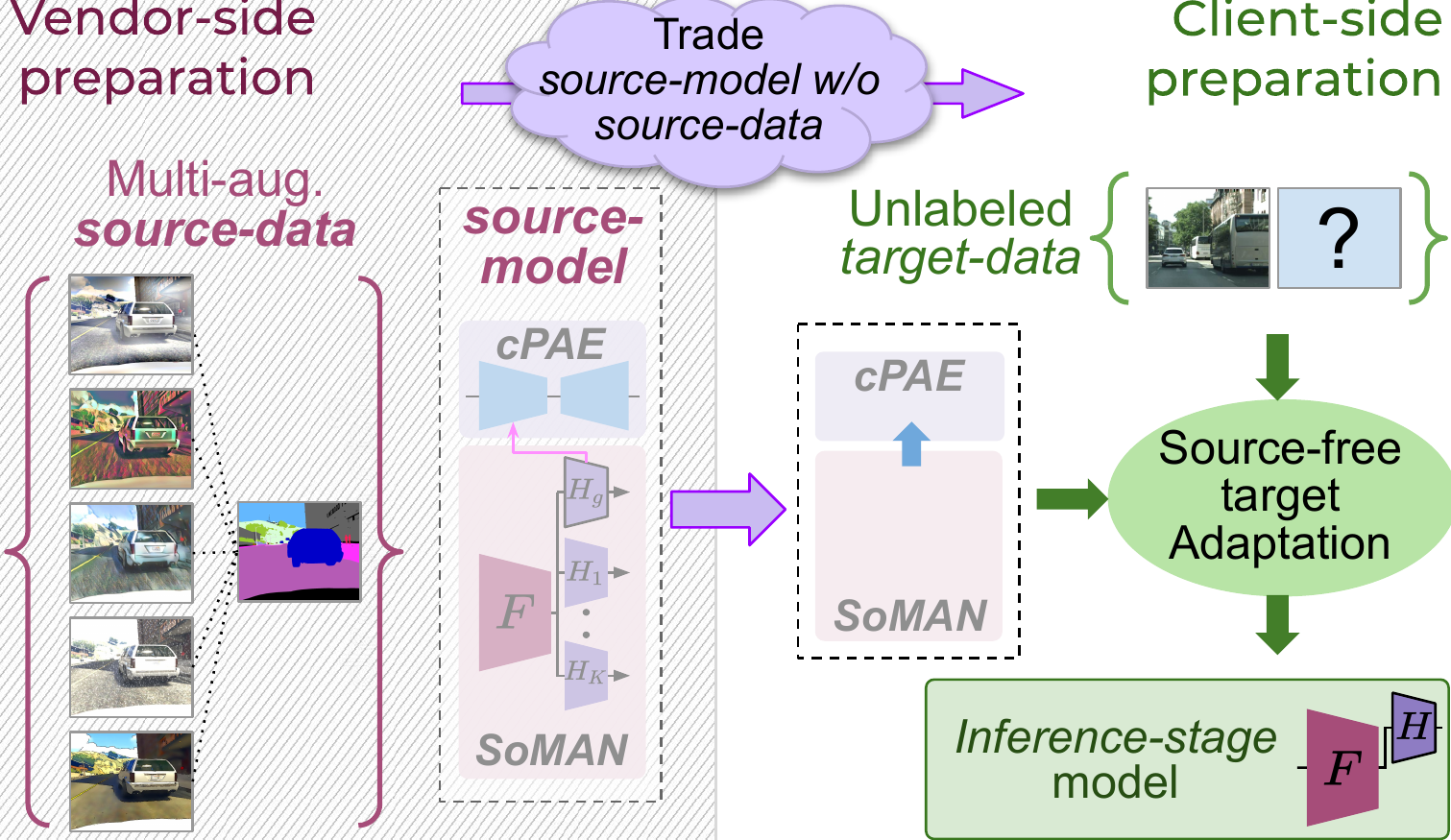}
    \caption{\small In source-free DA, the vendor accesses source-data to prepare a foresighted source-model. Following this, the client receives only the source-model to perform unsupervised target adaptation while prevented access to the proprietary source-data.
    }
    \vspace{-3mm}
    \label{fig:DASS_fig1}
\end{figure}

Unsupervised domain adaptation (DA) is one of the primary ways to address such problems.
Here, the goal is to transfer the knowledge from a labeled source domain to an unlabeled target domain. 
The major limitation of typical DA approaches \cite{saito2018maximum} is the requirement of concurrent access to both source and target domain samples.
While concurrent access better characterizes the distribution shift, it is a major bottleneck 
for real-world deployment scenarios.
Consider a modern corporate dealing where the \textit{vendor} organization has access to a large-scale labeled dataset (\ie \textit{source-data}) which is used to train a {\textit{source-model}}. 
The \textit{vendor} finds multiple \textit{clients} interested in deploying the \textit{source-model} in their specific target environments. However, both parties are restrained from data sharing due to proprietary, 
privacy, or profit related concerns. 
This motivates us to seek learning frameworks where the \textit{vendor} can trade only the \textit{source-model} and the client can perform target adaptation without the \textit{source-data}. This special case of domain adaptation \cite{li2020model, kundu2020universal, liang2020we} is \textit{Source-Free Domain Adaptation} (SFDA).

In this work, we aim to develop an SFDA framework for semantic segmentation of urban road scenes. In a co-operative setup, both vendor and the client must adopt specialized learning strategies to benefit the end goal. 

\noindent
\textbf{a) Vendor-side strategies.} These strategies can be discussed under two broad aspects viz. source dataset and training strategy. 
The vendor must acquire a substantially diverse large-scale dataset aiming to subsume unknown target scenarios. In literature, Multi-Source DA (MSDA) \cite{zhao2019multi, venkat2020msda, aggarwal2020wamda} and domain generalization (DG) \cite{li2017deeper} works use multiple labeled source domains to improve target generalization. However, gathering annotation for more than one domain is costly and time-consuming \cite{cordts2016cityscapes}.
Thus, we focus on developing a strategy to simulate multiple novel domains from samples from a single labeled domain.
Carefully crafted augmentations randomly perturb the task-irrelevant factors (such as stylization, texture modulation, \etc), facilitating the learning of domain-invariant representations.
Hence, we devise multiple augmentation-groups (\texttt{AG}s), where each group modulates the image by varying certain statistics thereby constructing virtual, labeled source-domains, to be treated as a multi-source dataset.

Next, we focus on developing an effective training strategy. 
The naive solution would be to train a single model on 
the entire multi-source dataset to 
learn domain-generic features. 
However, this can lead to sub-optimal performance if a certain \texttt{AG} alters the task-relevant causal factors \cite{ilse2021selecting}. 
Further, the target domain may be similar to one or a combination of \texttt{AG}s. 
In such cases, domain-specific (\texttt{AG}-specific) learning is more helpful. This motivates us to seek a domain-specific framework to complement the domain-generic model. Thus, we give theoretical insights to
analyze domain-specific hypotheses and propose \textit{Source-only Multi-Augmentation Network} (\texttt{SoMAN}) as shown in Fig. \ref{fig:DASS_fig1}.

Going forward, we recognize that \texttt{SoMAN} may lack the ability to capture inductive bias, which would prevent the model from manifesting structurally consistent predictions. 
This is particularly important for dense prediction tasks \cite{kundu2019umadapt,nath2018adadepth}. 
Modeling general context dependent priors 
encourages the prediction of plausible scene segments while discouraging common irregularities (\eg merged-region or split-region issues~\cite{kundu2020vrt}). To this end, we introduce a separate model namely, \textit{conditional Prior-enforcing Auto-Encoder} (\texttt{cPAE}). \texttt{cPAE} is trained on segmentation maps available with the vendor, and used at the client-side to improve the {source-free adaptation} performance. 

\noindent
\textbf{b) Client-side strategies.} 
We draw motivation from pseudo-label based self-training approaches \cite{grandvalet2005semi, zou2018unsupervised}. 
The target samples are passed through the \textit{source-model} to select a set of pseudo-labels which are later used to finetune the network.
In the absence of \textit{source-data}, effectiveness of such self-training depends on the following two aspects. First, the training must be regularized to retain the \textit{vendor-side}, \textit{task-specific knowledge}. We address this by allowing only a handful of weights to be updated while others are kept frozen from the \textit{vendor-side} training. Second, the pseudo-label selection criteria must overcome issues related to label-noise and information redundancy. We address this by selecting the optimal prediction from the \texttt{SoMAN}-heads and using the pruned output after forwarding through \texttt{cPAE}. 

In summary, we make the following main contributions:
\begin{itemize}

\vspace{-2mm}
\item We propose to address \textit{source-free} DA by casting the \textit{vendor-side} training as multi-source learning. To this end, we provide theoretical insights to analyze different ways to aggregate the domain-specific hypotheses. It turns out that a combination of domain-generic and \textit{leave-one-out} configuration performs the best.

\vspace{-2mm}
\item While accessing a single source domain, we propose a systematic way to select a minimal set of effective augmentations to resemble a multi-source scenario. The vendor uses this to develop a multi-head network, \texttt{SoMAN} subscribing to the leave-one-out configuration.

\vspace{-2mm}
\item Aiming to have a strong support for the spatially-structured segmentation task, we develop a conditional prior-enforcing auto-encoder. This encourages plausible dense predictions thereby enhancing the quality of pseudo-labels to aid the \textit{client-side} self-training.

\vspace{-2mm}
\item  Our \textit{source-free} framework achieves \textit{state-of-the-art} results for both GTA5 $\to$ Cityscapes and SYNTHIA $\to$ Cityscapes adaptation benchmarks, even when compared against the \textit{non-source-free} prior arts. 
\end{itemize}

\section{Related Work}
\label{sec:related_work}

\noindent 
Here, we briefly review the segmentation DA literature \cite{technologies8020035}.

\vspace{0.5mm}
\noindent \textbf{Feature-space DA.} 
The early works in DA for semantic segmentation are inspired from the GAN framework \cite{goodfellow2014generative}, involving training a segmentation network to confuse a domain discriminator enforcing domain invariance on the latent features \cite{hoffman2016fcns}. 
Several works \cite{chen2018road, li2019bidirectional, huang2018domain, hoffman2018cycada} utilized this discriminative alignment \cite{zhu2018penalizing, chen2019crdoco, Luo_2019_ICCV, du2019ssf} while adding complementary modules \cite{Haoran_2020_ECCV, dong2020cscl, wang2020differential} to improve adaptation. Another line of works \cite{sankaranarayanan2018learning, chen2019learning, luo2019taking, toldo2021unsupervised, yang2020label, yang2020adversarial} use the same framework on low-dimensional output space \cite{yu2021dast, kim2021crossdomain, yang2021context, tsai2019domain, vu2019advent, vu2019dada} instead of high-dimensional feature space. 
However, these works require cumbersome adversarial training and rely on source-target co-existence.

\vspace{0.5mm}
\noindent \textbf{Image-space DA.} The success of CycleGAN \cite{zhu2017unpaired} for image-to-image translation led to several DA approaches \cite{li2019bidirectional, hoffman2018cycada, chen2019crdoco, gong2019dlow, musto2020semantically} utilizing it for input-level adaptation while also addressing semantic consistency in the transformed images. Another category of works \cite{chang2019all, pizzati2020domain, zhang2018fully, choi2019self, wu2019ace} explore style-transfer techniques for input-level perceptual invariance~\cite{wu2018dcan, yang2020fda, luo2020ASM, lee2020unsupervised, wang2020consistency} between source and target domains. 
However, these works also assume the co-existence of source and target domains. 

\vspace{0.5mm}
\noindent \textbf{Source-free DA.} Bateson \etal \cite{bateson2020sourcerelaxed} perform \textit{source-free} DA for medical segmentation using entropy minimization and class-ratio alignment.
Concurrent \textit{source-free} works use data-free distillation, self-training, patch-level self-supervision \cite{liu2021source} and feature corruption with entropy regularization \cite{sivaprasad2021uncertainty} focused on target adaptation.
In contrast, we develop a novel approach for vendor-side source training. 

\vspace{0.5mm}
\noindent \textbf{DA via self-training.} 
Early works \cite{zou2018unsupervised, li2019bidirectional, zou2019confidence} use highly confident target predictions as pseudo-labels, selected using a confidence threshold. 
To improve the pseudo-labels, prior works used prediction ensembling \cite{chen2019domain, yang2020fda, zheng2019unsupervised, zheng2020unsupervised}, extra networks \cite{choi2019self}, applied constraints \cite{subhani2020learning}, modified the confidence thresholding technique \cite{mei2020instance, li2020content, shin2020two-phase}, utilized image-level pseudo-labels \cite{Paul_WeakSegDA_ECCV20} and intra-domain (easy-hard) adversarial training \cite{Pan_2020_CVPR}.
Most prior arts use labeled source with self-training to retain \textit{task-specific} source knowledge. 

\noindent \textbf{DG and MSDA.} 
\cite{zhao2019multi} use multiple synthetic datasets for Multi-Source DA (MSDA) in segmentation.
Restricted to a single source setting, we use data augmentation techniques to generate new domains.
In the presence of multi-source data, the vendor-side training is equivalent to domain generalization \cite{yue2019domain, chen2020automated, pan2018IBN} as it does not involve training on target.

\section{Approach}

Consider a set of source image and segmentation pairs $(x_s,y_s)\in\mathcal{D}_s$ where the source images $x_s$ are drawn from a marginal distribution $p_s$. The unlabeled target images $x_t \in \mathcal{D}_t$ are drawn from $p_t$. However, the output segmentation maps follow a single marginal distribution $p_y$. The goal is to learn a mapping $\hat{y}_s=h(x_s)$ 
that can generalize well for $x_t$. 
The proposed \textit{source-free} domain adaptation is broadly divided into two: \textit{vendor-side} and \textit{client-side}.

\subsection{Vendor-side Strategy}

In the absence of target data, the vendor's task effectively reduces to domain generalization (DG) \cite{li2017deeper}. DG is shown to be highly effective in the presence of multiple source domains. Thus, we plan to cast the vendor-side model preparation as a multi-source representation learning problem. 

\begin{figure*}
    \centering
    \vspace{-4mm}
    \includegraphics[width=\textwidth]{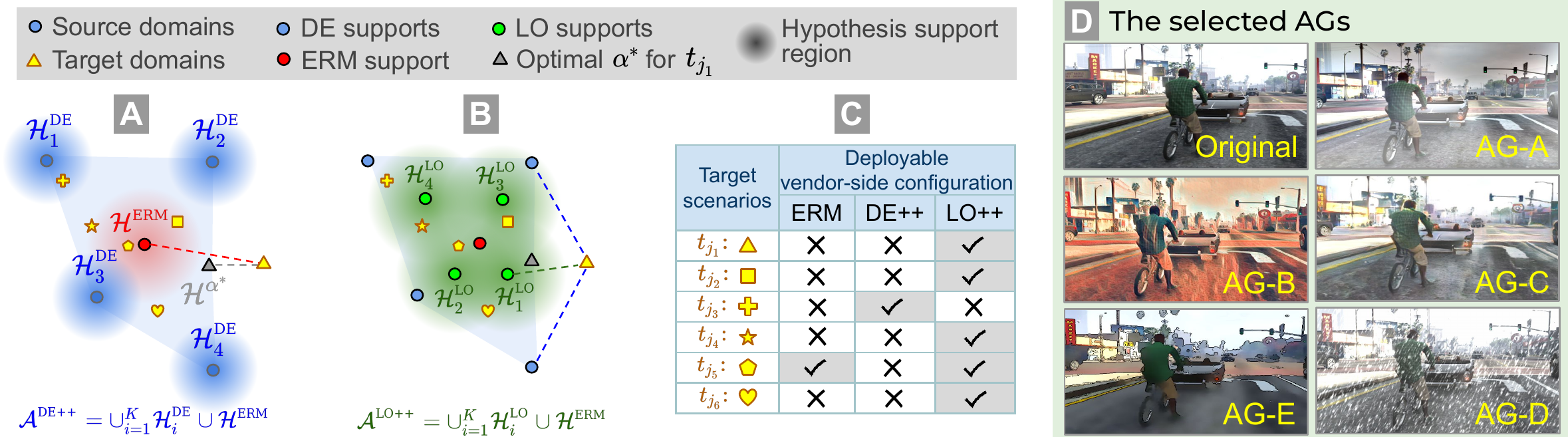}
    \vspace{-4mm}
    \caption{An illustration of hypothesis subspace constituents of \textbf{A.} $\mathcal{A}^\text{DE++}$ and \textbf{B.} $\mathcal{A}^\text{LO++}$. The positions of yellow signs represent the best non-source-free hypothesis $h^*_j$ for different target domains $t_j$. In the source-free paradigm, for each target $t_j$, the closest vendor hypothesis constituent would provide the support for a reasonable adaptation. \textbf{C.} The tick and cross marks for different target scenarios (the rows) denote suitability of the corresponding vendor-side configurations (the columns). For example, $\mathcal{H}^\text{ERM}$ is the best support for $t_{j_5}$, $\mathcal{H}^\text{DE}_1$ is the best support for $t_{j_3}$, $\mathcal{H}^\text{LO}_3$ is the best support for $t_{j_2}$, \etc Note that, \textit{LO++} is equipped to reasonably support a wide range of target scenarios.
    \textbf{D.} Visual illustration of selected \texttt{AG}s (augmented domains) based on the proposed augmentation selection criteria.
    }
    \vspace{-2mm}
    \label{fig:DASS_fig2}
\end{figure*}

\vspace{1mm}
\noindent \textbf{Non-source-free paradigm.} We assume access to $K$ source datasets $(x_{s_i}, y_{s_i}) \in \mathcal{D}_{s_i} \; \forall \; i \in [K]\!= \!\{1, 2, \dots, K\}$ where images $x_{s_i}$ are drawn from marginal distribution $p_{s_i}$. In non-source-free paradigm, the objective is to utilize all the domains (including the target) to realize a hypothesis $h^* = \argmin_{h \in \mathcal{A}} \epsilon_t(h)$ with a small target error, where

\vspace{-5mm}
\begin{equation}
\label{eqn:optimal_target_error}
    \epsilon_t(h) = \expectation_{(x, y)\sim p_t} [\mathcal{L}(h(x), y)]
    \;\; \text{where} \;\; h \in \mathcal{H}^{\alpha^*}\! \subset\! \mathcal{A}
\end{equation}
\vspace{-2mm}

\noindent Here, $\mathcal{L}$ is the loss and $\mathcal{A}$ is the hypothesis space. 
$\mathcal{H}^{\alpha^*}\subset \mathcal{A}$ can be interpreted as a hypothesis subspace spanning the hypotheses that can be learned using the best convex combination of sources $\alpha^* \in \Delta = \{\alpha \in [0, 1]^K:\sum_{i=1}^K\alpha[i]=1\}$ in the presence of concurrent access to $\{\mathcal{D}_{s_i}\}_{i=1}^K$ and $\mathcal{D}_t$, \ie $\alpha^* = \argmin_\alpha(\argmin_{h \in \mathcal{A}^\alpha} \epsilon_t(h))$.

While operating in a source-free paradigm \cite{kundu2020towards, kundu2020class}, let the vendor be approached by $M$ number of clients, each with different target domains 
$t_j \; \forall \; j \in [M]$. For every target $t_j$, there exists a specific $\alpha_j^*$ such that $\epsilon_{t_j}(h \in \mathcal{H}^{\alpha_j^*}) \leq \epsilon_{t_j}(h \in \mathcal{H}^\alpha) \; \forall \; \alpha \in \Delta$. However, in the absence of concurrent access to source and target domains (SFDA), it is not possible to optimize for $\alpha_j^*$ for any target $t_j$. 
Thus, we propose a source-free multi-domain paradigm.

\vspace{1.5mm}
\noindent \textbf{Definition 1. (Source-free multi-domain paradigm)} \textit{Consider a vendor who has access to labeled data $\{\mathcal{D}_{s_i}\}_{i=1}^K$ from $K$ source domains and a client who has access to unlabeled target data $\mathcal{D}_{t_j}$. In the source-free paradigm, the vendor prepares a prescient model with an immutable hypothesis support set $\mathcal{A}^\text{SF}$ (a union of certain hypothesis supports) 
without any information about $t_j$. This model is traded with the client for target adaptation without any data sharing.}

In the hypothetical scenario of source-target concurrent access, the client can determine the best $\alpha_j^*$ such that $\epsilon_{t_j}(h \in \mathcal{H}^{\alpha_j^*})\leq\epsilon_{t_j}(h\in\mathcal{A}^\textit{SF})$. The proposed paradigm not only enables adaptation without any data sharing, but also enables the vendor to prepare a single \textit{source-model} for all future clients. Thus, the process becomes more efficient for both vendor and client in terms of compute and storage.

\vspace{-5mm}
\subsubsection{Multi-source representation learning}
\vspace{-2mm}

Under \textit{source-free}, the vendor's objective would be to realize a learning setup that would generalize to a wide range of unseen targets. While aiming to learn a single hypothesis, empirical risk minimization (ERM) \cite{vapnik1992principles} would be the best solution (all domains weighted equally). Consider a scenario where $p_{t_j}$, \ie marginal distribution of the target $t_j$, matches with the marginal of one of the source domains. Here, the domain-specific expert for that source domain would definitely outperform the ERM baseline. To this end, a hypothesis support set $\mathcal{A}^\textit{SF}$, \ie a union of certain hypothesis supports, would provide better flexibility for SFDA. With this intent, we discuss the following configurations.

\vspace{1mm}
\noindent \textbf{a) ERM.} 
Under ERM configuration, we set $\mathcal{A}^\textit{SF}=\mathcal{H}^\text{ERM}$ where $\mathcal{H}^\text{ERM}$ is formed with equal weightage to all the multi-source domains \ie $\alpha[i] = \frac{1}{K} \, \forall \, i \in [K]$.

\vspace{1mm}
\noindent \textbf{b) Domain-experts++ (DE++).} 
This configuration encompasses a set of $K+1$ hypothesis supports. This includes $K$ number of domain-specific experts alongside one ERM support. Thus, we set $\mathcal{A}^\textit{SF}$ as $\mathcal{A}^{\text{DE++}} = \cup_{i=1}^K \mathcal{H}^{\text{DE}}_i \cup \mathcal{H}^{\text{ERM}}$. 
For $i^\text{th}$ support $\mathcal{H}^{\text{DE}}_i$, $\alpha_i[i'] = \mathbbm{1}_{i^\prime=i} \; \forall \, i' \in [K]$ where $\mathbbm{1}$ is the indicator function (1 if input condition is true, else 0).

\vspace{1mm}
\noindent \textbf{c) Leave-one-out++ (LO++).} 
It may happen that using a particular source may cause information loss that hinders optimal adaptation for a future target. To improve support for such targets, we introduce leave-one-out (LO) hypothesis support where $i^\text{th}$ subspace $\mathcal{H}_i^\text{LO}$ is formed by leaving one domain out, \ie with $\alpha_i[i'] = \frac{1}{K-1}\mathbbm{1}_{i \neq i'} \, \forall \, i' \in [K]$. Similar to \textit{DE++}, \textit{LO++} also includes $K+1$ hypothesis supports, \ie $K$ number of LO supports with one ERM. 
Thus, we set $\mathcal{A}^\textit{SF}$ as $\mathcal{A}^{\text{LO++}} = \cup_{i=1}^K \mathcal{H}^{\text{LO}}_i \cup \mathcal{H}^{\text{ERM}}$.

We include the ERM support, \ie $\mathcal{H}^\text{ERM}$, in both \textit{LO++} and \textit{DE++} to provide complementary domain-generic information alongside the different forms of domain-specific 
information. Here, the individual hypothesis supports are implemented as separate classifier heads trained on a common feature extractor 
(Sec \ref{subsubsec:vendorsidearch}).
Note that we only consider options that require $K$ heads while other domain-specific solutions like leave-$r$-out have higher computational cost requiring $\binom{K}{r}$ heads. Next, we discuss a result comparing the target error $\epsilon_t(h)$ of the three configurations.

\vspace{1mm}
\noindent \textbf{Result 1.} 
\textit{Consider DE++ hypothesis space} $\mathcal{A}^{\text{DE++}}$, \textit{LO++ hypothesis space} $\mathcal{A}^{\text{LO++}}$,
\textit{and unseen target data $\mathcal{D}_t$. Then,}
\vspace{-2mm}
\begin{equation}
\label{eqn:DE-LO-result}
    \begin{split}
    \epsilon_t(h \in \mathcal{A}^\text{LO++}) \leq \epsilon_t(h \in \mathcal{H}^\text{ERM}) \\
    \epsilon_t(h \in \mathcal{A}^\text{DE++}) \leq \epsilon_t(h \in \mathcal{H}^\text{ERM})
    \end{split}
\end{equation}

As depicted in Fig. \ref{fig:DASS_fig2}, the distributed subspace constituents of $\mathcal{A}^\text{DE++}$ and $\mathcal{A}^\text{LO++}$ provides better support for a wide range of unknown target domains as compared to the same by $\mathcal{H}^\text{ERM}$. Thus, in Eq. \ref{eqn:DE-LO-result}, $\epsilon_t(h \in \mathcal{H}^\text{ERM})$ acts as an upper bound for the target risk, particularly in source-free paradigm. Also, the equality holds as both $\mathcal{A}^\text{DE++}$ and $\mathcal{A}^\text{LO++}$ already include $\mathcal{H}^\text{ERM}$ as a constituent subspace.

\vspace{1mm}
\noindent
\textbf{Comparison between DE++ and LO++.}
Though, both \textit{DE++} and \textit{LO++} are better alternatives over ERM, it is not possible to write a general inequality involving only the target errors for \textit{DE++} and \textit{LO++} configurations. Note that, as shown in Fig. \ref{fig:DASS_fig2}, for certain target scenarios, target error for DE++ would be less than the same for LO++ and vice versa. However, considering a reasonable domain-shift among the source domains, \textit{LO++} provides lower target error over \textit{DE++} for a wide range of practical target scenarios (see Fig. \ref{fig:DASS_fig2}\textcolor{red}{C}).
\textit{DE++} wins particularly for cases when $p_t\approx p_{s_{i^\prime}}$ for $i^\prime\!\in\![K]$ which is generally quite rare. \textit{LO++} wins for a wide range of unique target scenarios.

\begin{figure*}
    \centering
    \vspace{-3mm}
    \includegraphics[width=\textwidth]{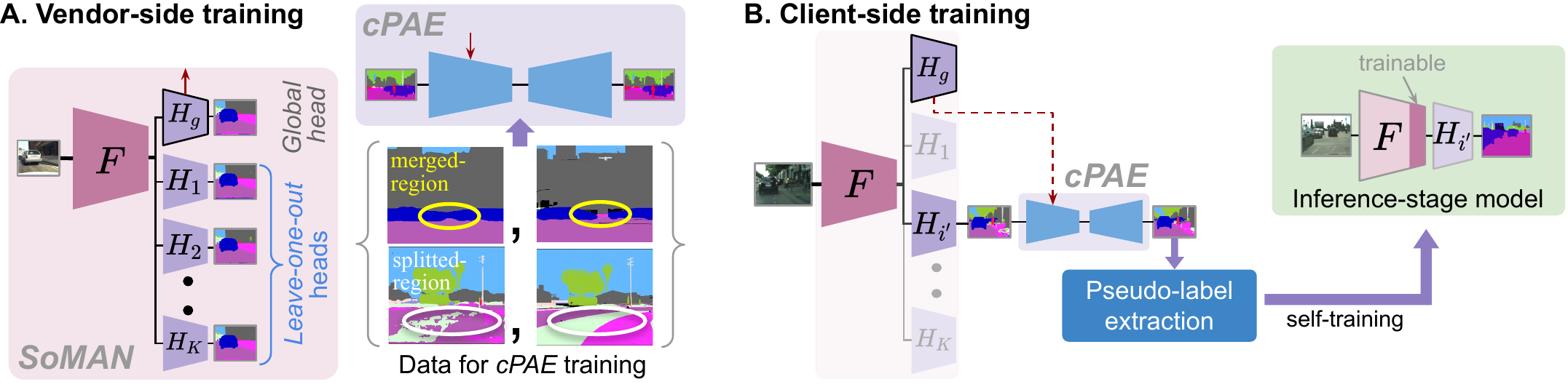}
    \vspace{-4.5mm}
    \caption{\textbf{A.} \texttt{SoMAN} constitutes of a global-head with multiple leave-one-out heads (left). Vendor also trains \texttt{cPAE} to discourage prediction irregularities. \textbf{B.} 
    Client receives \texttt{SoMAN} and \texttt{cPAE} from vendor, and extracts robust and meaningful pseudo-labels for target samples via the optimal head $H_{i'}$ to perform \textit{source-free} DA. The inference model uses only the optimal head $H_{i'}$ (no computational overhead).
    }
    \vspace{-2mm}
    \label{fig:DASS_fig3}
\end{figure*}

\vspace{-4mm}
\subsubsection{Preparing virtual multi-source domains}
\vspace{-2mm}

\noindent Having identified \textit{LO++} as the best option, we focus on obtaining the multi-source data. 
Though, we intend to expand our source-data horizon, we are restricted to a single labeled source domain.
Thus, we plan to use diverse data augmentations to simulate a multi-source scenario.

\vspace{1mm}
\noindent \textbf{Characterizing multi-domain data.}
Consider a hypothetical data generation process \cite{piratla20efficient} for the source domain: A data generator $\phi$ uses the causal class factor $f_y$ and the non-causal domain-related factor $f_s$ to construct a data sample $x_s = \phi(f_y, f_s)$. Next, a set of domain-varying class-preserving augmentations $\{\mathcal{T}_i\}_{i=1}^K$ are applied to obtain,

\vspace{-4mm}
\begin{equation}
    \label{eqn:aug_theory}
    x_{s_i} = \mathcal{T}_i(x_s) = \phi(f_y, f_i + \gamma_i f_s);\;\;\; \gamma_i \in \mathbb{R}
\end{equation}

Here, $\mathcal{T}_i$ modifies the original domain-specific factor $f_s$ by a weight $\gamma_i$ (without altering $f_y$) and introduces a new augmentation related domain-specific factor $f_i$. 
Thus, the augmentations modify the non-causal factors to simulate novel domains. 
The augmented datasets are realized by pairing the input with the corresponding label and are represented as $(x_{s_i} = \mathcal{T}_i(x_s), y_s)\in\mathcal{D}_{s_i}$ $\forall\, i\in[K]$. 

\vspace{1mm}
\noindent \textbf{Effect of number of source domains $K$.}
Having a very high $K$ would lead to significant overlap of the the \textit{leave-one-out} subspaces with the ERM, \ie nullify the advantage of \textit{LO++}. 
Further, a high $K$ would induce a higher computational complexity. Thus, it becomes essential to filter out augmentations through a principled selection criteria.

\vspace{1.5mm}
\noindent \textbf{Definition 2 (Augmentation selection criteria)} 
\textit{Using Eq.~\ref{eqn:aug_theory}, an augmentation $\mathcal{T}_i$ will be selected if 
$|\gamma_i| < 1$. 
We give a tractable surrogate for this condition,
using a hypothesis $h_s = \argmin_{h \in \mathcal{A}}\hat{\epsilon}_{(x, y)\in\mathcal{D}_s}(h)$ trained only on $\mathcal{D}_s$,}

\vspace{-2mm}
\begin{equation}
    \label{eqn:aug_selection}
    \hat{\epsilon}_{(\mathcal{T}_i(x), y)\in\mathcal{D}_{s_i}}(h_s) - \hat{\epsilon}_{(x, y)\in\mathcal{D}_s}(h_s) > \tau; \;\;\; 
\end{equation}

\noindent \ie \textit{the gap between the empirical risks (}\ie \textit{$\hat{\epsilon}$) of $h_s$ on $\mathcal{D}_{s_i}$ and $\mathcal{D}_s$ should be greater than a threshold $\tau$. This ensures that $\mathcal{T}_i$ exerts a substantial alteration in the image statistics equivalent to the style gap between two diverse domains.}

\vspace{1mm}
Intuitively, an augmentation is selected if it can suppress (\ie $|\gamma_i| \!<\! 1$) the original domain factor $f_s$.
In practice, $\gamma_i$ is intractable in the absence of disentangled $f_y$ and $f_s$. Thus, we rely on Eq. \ref{eqn:aug_selection} whose LHS expresses the generalization error due to the domain-specific bias (\ie the correlation between $f_s$ and $y_s$) inculcated in $h_s$.
See Suppl. for an extended explanation of the selection criteria.
These diverse domains will help the model generalize to a wider range of targets.
Henceforth, we denote each of these as an \texttt{AG} (augmentation-group), each representing a specific type of class-preserving, domain-varying augmentation.

\vspace{-4mm}
\subsubsection{Vendor-side architecture and training}
\label{subsubsec:vendorsidearch}
\vspace{-1.5mm}

\noindent \textbf{Architecture.} Considering the advantages of \textit{LO++},
we propose a Source-only Multi-Augmentation Network, \texttt{SoMAN}, which is essentially a multi-head architecture with a shared CNN backbone $F$ (see Fig. \ref{fig:DASS_fig3}\textcolor{red}{A}). Along with a global output head $H_g$ which is optimized using ERM, we employ leave-one-out heads $\{H_i\}_{i=1}^K$ trained to be sensitive towards the corresponding \texttt{AG} (\ie $\mathcal{T}_i$) while being invariant to others. Formally, the global head is trained using all the augmented datasets \ie $\mathcal{D}_{s_g}=\cup_{i=1}^K \mathcal{D}_{s_i}$ and each non-global head $H_i$ is trained using a head-specific dataset $\mathcal{D}_{s_i}^- = \mathcal{D}_{s_g}\setminus\mathcal{D}_{s_i}$.

\noindent \textbf{Training procedure.} The \texttt{SoMAN} architecture is trained by simultaneously optimizing the spatial segmentation losses computed at the end of each output head. This encourages $F$ to extract a rich multi-source representation which retains domain-sensitive cues (as a result of the leave-one-out setup) alongside the extraction of domain-generic features. We denote the output of global head as  $h_g=H_g(F(x))$. Following a similar convention, output of the leave-one-out heads are denoted by $h_i = H_i(F(x))$. Thus, the final objective for end-to-end training of \texttt{SoMAN} is formulated as,

\vspace{-6mm}
\begin{equation}
    \label{eqn:soman_objective}
    \min_{\theta} \sum_{i=1}^K \expectation_{(x, y) \in \mathcal{D}_{s_i}^-} [-\langle y, \log h_i  \rangle ]+ \expectation_{(x, y)\in \mathcal{D}_{s_g}} [-\langle y, \log h_g  \rangle ]
\end{equation}

\noindent Here, $\theta$ denotes a set of parameters from all the heads, \ie 
$\theta_F$, 
$\theta_{H_g}$, $\{\theta_{H_i}\}_{i=1}^K$ while $\langle ., . \rangle$ represents the dot product of the two inputs. In practice, the expectations are computed by sampling mini-batches from the corresponding datasets.

\vspace{-4mm}
\subsubsection{Conditional prior-enforcing autoencoder (cPAE)}
\vspace{-1.5mm}

In dense prediction tasks such as semantic segmentation, not all predictions are equally likely. Though the target annotations are not available during the client-side training, we aim to explicitly impart the general knowledge of scene prior to constrain the solution space. The use of scene prior would encourage plausible scene segments while discouraging irregularities (see Fig. \ref{fig:DASS_fig3}{\color{red}A}) such as ``car flying in the sky", ``grass on road", ``split car shape", ``merged pedestrians", etc. We recognize that the \texttt{SoMAN} may lack the ability to capture the above discussed inductive bias.

\vspace{0.5mm}
\noindent \textbf{How can structural inductive bias be captured?}
We propose a conditional Prior-enforcing Auto-Encoder (\texttt{cPAE}), denoted by $Q$, that refines the predicted segmentation maps (seg-maps) conditioned on domain-generic features extracted from \texttt{SoMAN}. Instead of training it as a plain auto-encoder, we plan to train it as a denoising auto-encoder. The question that arises here is: how do we simulate noise for the \texttt{cPAE} inputs? We take advantage of sensitivity of leave-one-out heads to the corresponding \texttt{AG}s to simulate noisy seg-maps. Thus, the \texttt{cPAE} output distribution is $Q(y|F_g(x_{s_i}), \hat{y})$ where $\hat{y}=H_i(F(x_{s_i}))$. $F_g$ consists of the backbone F and the first block of $H_g$ such that $F_g(x_{s_i})$ are domain-generic features since $H_g$ is trained using all \texttt{AG}s. We train the \texttt{cPAE} to align its output distribution with the true source label distribution $p_s$ as follows

\vspace{-5mm}
\begin{equation}
    \label{eqn:cpae_objective}
    \min_{\theta_Q} \sum_{i=1}^K\mathbb{E}_{(x, y)\in \mathcal{D}_{s_i}}[\text{KL}(p_s(y), Q(y|F_g(x), \hat{y}))]
\end{equation}

\noindent Here, $\text{KL}$ indicates the Kullback-Leibler divergence. In practice, cross-entropy loss between the \texttt{cPAE} output and ground truth seg-map is used, derived from the KL term.

\subsection{Client-side Strategy}
\noindent Since the client can access only unlabeled target data $x^t \in \mathcal{D}^t$, we propose the use of self-training for this source-free adaptation step. However, this presents two caveats, 

\vspace{1mm}
\noindent \textbf{a) Risk of overfitting to wrong overconfident predictions.} 
To counter this, we propose to utilize the multiple heads of \texttt{SoMAN} and the \texttt{cPAE} to generate reliable pseudo-labels.

\vspace{1mm}
\noindent \textbf{b) Loss of task-relevant information.} To avoid this, we aim to preserve the task-specific knowledge of the vendor model. While prior arts trained the entire model,
we propose to train only a handful of weights belonging to the later layers of $F$ while others are frozen from vendor-side. 
The frozen output heads hold useful, domain-generic, task-related inductive bias.
It also constrains the optimization to operate within the hypothesis subspace of the vendor-side initialization. Thus, the client can leverage the vendor's foresighted preparation to avoid sub-optimal solutions.

\vspace{-4mm}
\subsubsection{Pseudo-label extraction via cPAE}
\vspace{-1.5mm}

\noindent Since pseudo-labels are the only supervision signal in the proposed \textit{source-free} self-training, it is crucial to ensure that they are highly informative and reliable. To this end, we propose to utilize the optimal head of the vendor provided \texttt{SoMAN} and the \texttt{cPAE} to obtain improved pseudo-labels. We consider the optimal head as the one that produces the lowest average self-entropy for the target training dataset. Formally, $H_{i^\prime}$ is the optimal head where $i^\prime=\argmin_{i \in \{g, [K]\}}\sum_{x \in \mathcal{D}_t}\{-\langle h_i, \log h_i \rangle \}$ where $h_i = H_i(F(x))$. The optimal prediction can be represented as $Q(h_{i^\prime})$. Note that we denote the \texttt{cPAE} output as $Q(h_{i^\prime})$ omitting the conditional feature input for simplicity. 

Using the optimal prediction, we follow \cite{li2019bidirectional, zou2018unsupervised} for the confidence thresholding method. Particularly, we choose the top 33\% of the most confident pixel-level predictions per class over the entire target training set. This gives a target pseudo-labeled subset $(x_t, \hat{y}_t) \in \hat{\mathcal{D}}_t $ for self-training. Note that, the unselected pixels are assigned a separate, `unknown' class which is not considered in training.

\vspace{-4mm}
\subsubsection{Source-free adaptation via self-training}
\vspace{-1.5mm}

\noindent
We perform three rounds of self-training, following \cite{yang2020fda}, where each round consists of pseudo-label extraction in an offline manner followed by supervised training on the extracted pseudo-labels. Entropy minimization is used as a regularizer during self-training.
Further, we use the shared backbone $F$ along with the optimal head, $H_{i^\prime}$, for both self-training and test-time inference. 
Formally, 

\vspace{-3mm}
\begin{equation}
    \label{eqn:self_training_objective}
    \min_{\theta_F} \expectation_{(x_t, \hat{y}_t) \in \hat{\mathcal{D}}_t} [-\langle \hat{y}_t, \log H_{i^\prime}(F(x_t)) \rangle ]
\end{equation}
\vspace{-1mm}

\vspace{-5mm}
\subsubsection{Test-time inference}
\vspace{-2mm}

\noindent
As we propose only optimal head (\ie $H_{i^\prime}$) self-training, our inference-stage model is $H_{i^\prime}(F(x_t))$ as shown in Fig.~\ref{fig:DASS_fig3}\textcolor{red}{B}. However, \texttt{cPAE} provides a further improvement in performance if used during inference. But, unless otherwise specified, the experiments use only $H_{i^\prime}(F(x_t))$ for self-training and evaluation, for a fair comparison. Note that, `w/ \texttt{cPAE}' means that \texttt{cPAE} was used only for pseudo-label extraction.


\section{Experiments}
\label{sec:experiments}

We perform a thorough evaluation of our approach against \textit{state-of-the-art} prior works across multiple settings. 

\subsection{Experimental Settings}
\label{sec:expt_settings}
\noindent \textbf{a) Network architectures.} Following \cite{li2019bidirectional, yang2020fda}, we employ 2 widely-used network architectures for the DA setting on semantic segmentation, DeepLabv2 \cite{chen2017deeplab} with ResNet101 \cite{he2016deep_resnet} backbone and FCN8s \cite{long2015fully} with VGG16 \cite{simonyan2014very} backbone. 
See Suppl. for the complete details.

\vspace{1mm}
\noindent \textbf{b) Datasets.} We extensively evaluate the proposed approach on two popular synthetic-to-real benchmarks \ie GTA5$\to$Cityscapes and SYNTHIA$\to$Cityscapes. 
We provide the complete implementation details in the Suppl.

\vspace{1mm}
\noindent \textbf{c) Evaluation metric.} Following \cite{li2019bidirectional, yang2020fda}, we compute {per-class} IoUs as well as mean IoU (mIoU) over all 19 classes for the GTA5$\to$Cityscapes task. For SYNTHIA$\to$Cityscapes, we report the same for 13 and 16 classes because of the lower number of overlapping classes.
Following \cite{mei2020instance, Haoran_2020_ECCV, zhang2020transferring}, we use multi-scale testing.
Due to space limitations, we report mean IoUs for \textit{class-groups}\footnote{Background (\textbf{BG}) - building, wall, fence, vegetation, terrain, sky; Minority Class (\textbf{MC}) - rider, train, motorcycle, bicycle; Road Infrastructure Vertical (\textbf{RIV}) - pole, traffic light, traffic sign; Road Infrastructure Ground (\textbf{RIG}) - road, sidewalk; and Dynamic Stuff (\textbf{DS}) - person, car, truck, bus.} instead of reporting IoUs for each individual class.

\vspace{1mm}
\noindent \textbf{d) Augmentations.} 
We select the following $K=5$ \texttt{AG}s (see Fig. \ref{fig:DASS_fig2}\textcolor{red}{D}) using Definition \textcolor{red}{2} with the mIoU metric.

\textbf{Aug-A} (FDA \cite{yang2020fda}): 
This uses Fourier transform to transfer style from a reference image while retaining the semantic features~\cite{yang2020phase} of the input. While FDA~\cite{yang2020fda} transfers the style from target images, we do not access target data for vendor-side training. We use a small subset from style transfer dataset~\cite{huang2017adain} and random noise as reference images.

\textbf{Aug-B} (Style augmentation \cite{jackson2019style}): This technique uses a deep style transfer network for style randomization by randomly sampling a style embedding from a multivariate normal distribution instead of using reference style image. This provides practically infinite number of stylization options.

\textbf{Aug-C} (AdaIN \cite{huang2017adain}): This uses Adaptive Instance Normalization (AdaIN) layers
to inject style from a given reference image. 
In contrast to Aug-B, this provides a way to stylize images using a desired style image. 
We use a small subset from style transfer dataset \cite{huang2017adain} as reference images.

\textbf{Aug-D} (Weather augmentation) \cite{imgaug, michaelis2019dragon}: 
We use realistic weather augmentations to generate varying levels of snow and frost in the images. Compared to other \texttt{AG}s, this simulates realistic variations in the road scene images.

\textbf{Aug-E} (Cartoon augmentation) \cite{imgaug}: This technique generates cartoonized versions of input images. This augmentation is diverse and useful as it produces almost texture-less images as in cartoons or comic books.

\begin{table}[t]
\centering
\vspace{-3mm}
\caption{Quantitative evaluation on GTA5$\to$Cityscapes. Performance on different segmentation architectures: A (DeepLabv2 ResNet-101), B (FCN8s VGG-16). SF indicates \textit{source-free} adaptation. See Suppl. for the extended table with per-class IoUs. \textit{Ours (V)} indicates use of our vendor-side \texttt{AG}s with prior art, * indicates results produced using the released code of prior arts.}
\label{tab:gta2city}
\setlength{\tabcolsep}{4pt}
 \resizebox{1\columnwidth}{!}{
\begin{tabular}{llcccccccc}
    \toprule
    \# & Method & Arch. & SF & BG & MC & RIV & RIG & DS & mIoU \\ 
    \midrule
    1. & PLCA \cite{kang2020pixel} & A & $\times$ & 57.3 & 28.3 & 31.1 & 57.2 & 60.2 & 47.7 \\
    2. & CrCDA \cite{huang2020contextualrelation} & A & $\times$ & 57.5 & 24.5 & 33.8 & 73.9 & 57.6 & 48.6 \\
    3. & RPT \cite{zhang2020transferring} & A & $\times$ & 62.5 & 34.9 & 42.0 & 67.3 & 59.4 & 53.2 \\
    4. & DACS \cite{tranheden2021dacs} & A & $\times$ & 63.1 & 24.2 & 45.9 & 64.7 & 61.8 & 52.1 \\
    5. & FADA \cite{Haoran_2020_ECCV} & A & $\times$ & 61.9 & 26.7 & 35.0 & 70.8 & 56.7 & 50.1 \\
    6. & IAST \cite{mei2020instance} & A & $\times$ & 60.4 & 32.6 & 34.1 & 76.5 & 60.7 & 52.2 \\
    7. & \textit{Ours (V)} + FADA* & A & $\times$ & 62.8 & 27.1 & 35.3 & 71.1 & 57.2 & 50.6 \\
    8. & \textit{Ours (V)} + IAST* & A & $\times$ & 61.0 & 33.1 & 34.6 & 77.1 & 61.2 & 52.8 \\
    \arrayrulecolor{gray}\hline\arrayrulecolor{black}
    9. & URMA \cite{sivaprasad2021uncertainty} & A & \checkmark & 55.8 & 23.8 & 22.3 & 73.7 & 52.8 & 45.1 \\
    10. & SRDA* \cite{bateson2020sourcerelaxed} & A & \checkmark & 57.1 & 20.2 & 33.5 & 68.8 & 51.9 & 45.8 \\
    \rowcolor{gray!10} 11. & \textit{Ours (w/o \texttt{cPAE})} & A & \checkmark & 61.8 & 30.3 & 35.1 & 69.2 & 60.8 & 51.6 \\
    \rowcolor{gray!10} 12. & \textit{Ours (w/ \texttt{cPAE})} & A & \checkmark & 62.8 & 33.4 & 36.2 & 72.0 & 66.4 & \textbf{53.4} \\
    \midrule
    13. & LTIR \cite{kim2020learning} & B & $\times$ & 58.6 & 14.0 & 26.5 & 73.5 & 42.5 & 42.3 \\
    14. & FADA \cite{Haoran_2020_ECCV} & B & $\times$ & 57.7 & 16.3 & 25.8 & 71.7 & 50.1 & 43.8 \\
    15. & PCEDA \cite{yang2020phase} & B & $\times$ & 56.4 & 20.5 & 31.2 & 67.5 & 49.5 & 44.6 \\
    \arrayrulecolor{gray}\hline\arrayrulecolor{black}
    16. & SFDA \cite{liu2021source} & B & \checkmark & 51.6 & 7.8 & 15.9 & 58.6 & 43.7 & 35.8 \\
    \rowcolor{gray!10} 17. & \textit{Ours (w/o \texttt{cPAE})} & B & \checkmark & 54.7 & 19.9 & 27.3 & 66.2 & 50.3 & 43.4 \\
    \rowcolor{gray!10} 18. & \textit{Ours (w/ \texttt{cPAE})} & B & \checkmark & 49.9 & 30.3 & 32.9 & 74.9 & 50.8 & \textbf{45.9} \\
    \bottomrule
\end{tabular}}
\vspace{-1mm}
\end{table}


\begin{table}[t]
    \centering
    \caption{Ablation study for GTA5$\to$Cityscapes. * indicates 3 rounds of self-training after the mentioned method. The client-side ablations begin from the best vendor-side model.}
    \label{table:ablation}
    \setlength{\tabcolsep}{15pt}
     \resizebox{1\columnwidth}{!}{
    \begin{tabular}{llc}
    \toprule
     & Method & mIoU \\
    \midrule
    \multirow{4}{*}{Vendor-side} & Standard single-source* & 44.4 \\
    & Multi-source ERM* & 47.6 \\
    & Domain-experts++ (DE++)* & {48.0} \\
    & Leave-one-out++ (LO++)* & \textbf{51.6} \\
    \midrule
    \multirow{4}{*}{Client-side} & w/o \texttt{cPAE} & 51.6 \\
    & \hspace{15pt}+ Inference via \texttt{cPAE} & 52.5 \\
    & w/ \texttt{cPAE} & 53.4 \\
    & \hspace{15pt}+ Inference via \texttt{cPAE} & \textbf{54.2} \\
    \bottomrule
    \end{tabular}}
\vspace{-4mm}
\end{table}

\subsection{Discussion}

We provide an extensive ablation study of both the \textit{vendor-side} and the \textit{client-side} preparation.
Further, we show that our approach generalizes across novel target scenarios and is compatible to online domain adaptation.

\vspace{-5mm}
\subsubsection{Comparison with prior arts.}
\vspace{-2mm}

\begin{figure*}[t]
    \centering
    \vspace{-1mm}
    \includegraphics[width=\textwidth]{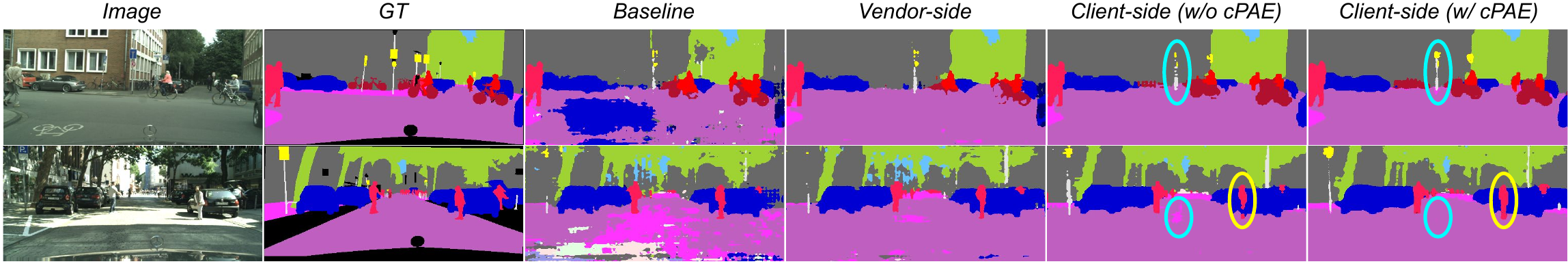}
    \caption{Qualitative evaluation of the proposed approach. Vendor-side model generalizes better than the baseline but performs worse than client-side due to the domain gap. Inculcating the prior knowledge from \texttt{cPAE} structurally regularizes the predictions and overcomes the merged-region (yellow circle) and split-region (blue circle) problems. See Suppl. for extended evaluation. \textit{Best viewed in color.}}
    \vspace{-1mm}
    \label{fig:DASS_fig5}
\end{figure*}

We compare our proposed approach with prior arts in Table \ref{tab:gta2city} and \ref{tab:synthia2city}. 
We also compare our vendor-side approach with prior DG works in Table \ref{tab:domaingeneralization}.
Our method achieves \textit{state-of-the-art} performance across all benchmarks. 
We also present the qualitative evaluation of our approach in Fig. \ref{fig:DASS_fig5}.

Our proposed client-side adaptation is more scalable compared to prior works like PCEDA \cite{yang2020phase}, RPT \cite{zhang2020transferring}, IAST \cite{mei2020instance} in two ways. First, our method does not require image-to-image translation networks (PCEDA) or adversarial training (RPT, IAST) thereby reducing the adaptation complexity. 
Also note that the frozen \texttt{cPAE} is used only to obtain better pseudo-labels and is not involved in backpropagation for adaptation training. Second, the client can perform adaptation to multiple different target domains without the complex vendor-side training and without access to the source data. We study the second aspect further in the paper. 

\vspace{1mm}
\noindent \textbf{a) Comparison with source-free prior arts.}
We implemented \cite{bateson2020sourcerelaxed} for GTA5$\to$Cityscapes (see \#10-12 in Table \ref{tab:gta2city}) since they only report results for single object segmentation. We outperform their approach by a significant margin (8.1\%). We also compare with concurrent source-free works \cite{liu2021source, sivaprasad2021uncertainty} (see \#9 vs. \#12, \#16 vs. \#18 in Table \ref{tab:gta2city} and \#7 vs. \#9 in Table \ref{tab:synthia2city}) and outperform them by $\sim$12\%.

\vspace{1mm}
\noindent \textbf{b) Disentangling the gains from use of augmented data.}

\noindent We show the results for 2 prior arts \cite{mei2020instance, Haoran_2020_ECCV} using our vendor-side \texttt{AG}s during training (\#5-8 in Table \ref{tab:gta2city}). While the performance improves compared to that originally reported, our proposed method (\#12 in Table \ref{tab:gta2city}) still outperforms them. Thus, the improvement of our proposed method depends not only on the use of \texttt{AG}s but also on the multi-head, leave-one-out \texttt{SoMAN} framework and the \texttt{cPAE}.

\begin{table}[t]
\centering
\vspace{-3mm}
\caption{Quantitative evaluation on SYNTHIA$\to$Cityscapes. Performance on different segmentation architectures: A (DeepLabv2 ResNet-101), B (FCN8s VGG-16). mIoU and mIoU* are averaged over 16 and 13 categories respectively. SF indicates whether the method supports \textit{source-free} adaptation. See Suppl. for the extended table with per-class IoUs.}
\label{tab:synthia2city}
\setlength{\tabcolsep}{2.85pt}
 \resizebox{1\columnwidth}{!}{
\begin{tabular}{llccccccccc}
    \toprule
    \# & Method & Arch. & SF & BG & MC & RIV & RIG & DS & mIoU & mIoU*\\ 
    \midrule
    1. & CAG \cite{zhang2019category} & A & $\times$ & 81.3 & 32.9 & 18.0 & 62.6 & 54.9 & 44.5 & 52.6 \\
    2. & USAMR \cite{zheng2019unsupervised}  & A & $\times$ & 81.3 & 33.0 & 25.1 & 60.7 & 61.7 & 46.5 & 53.8 \\
    3. & DACS \cite{tranheden2021dacs} & A & $\times$ & 85.4 & 38.1 & 23.3 & 52.8 & 63.1 & 48.7 & 54.8 \\
    4. & RPL \cite{zheng2020unsupervised} & A & $\times$ & 81.8 & 32.8 & 25.6 & 64.8 & 63.3 & - & 54.9 \\
    5. & IAST \cite{mei2020instance} & A & $\times$ & 83.9 & 38.9 & 29.9 & 61.7 & 63.4 & 49.8 & 57.0 \\ 
    6. & RPT \cite{zhang2020transferring} & A & $\times$ & 85.7 & 37.2 & 35.1 & 68.2 & 66.2 & 51.7 & 59.5 \\
    \arrayrulecolor{gray}\hline\arrayrulecolor{black}
    7. & URMA \cite{sivaprasad2021uncertainty} & A & \checkmark & 80.1 & 23.6 & 25.1 & 41.9 & 46.6 & 39.6 & 45.0 \\
    \rowcolor{gray!10} 8. & \textit{Ours (w/o \texttt{cPAE})} & A & \checkmark & 82.9 & 34.4 & 22.5 & 66.8 & 65.3 & 48.1 & 55.5 \\
    \rowcolor{gray!10} 9. & \textit{Ours (w/ \texttt{cPAE})} & A & \checkmark &84.3 & 42.2 & 29.3 & 69.8 & 67.8 & \textbf{52.0} & \textbf{60.1} \\
    \midrule
    10. & PyCDA \cite{lian2019constructing} & B & $\times$ & 75.4 & 16.4 & 24.0 & 53.6 & 47.6 & 35.9 & 42.6 \\
    11. & SD \cite{du2019ssf} & B & $\times$ & 79.2 & 6.3 & 10.7 & 64.4 & 54.4 & - & 43.4\\
    12. & FADA \cite{Haoran_2020_ECCV} & B & $\times$ & 82.1 & 16.1 & 15.1 & 58.2 & 52.6 & 39.5 & 46.0 \\
    13. & BDL \cite{li2019bidirectional} & B & $\times$ & 78.3 & 25.2 & 17.7 & 51.2 & 50.5 & 39.0 & 46.1 \\
    14. & PCEDA \cite{yang2020phase} & B & $\times$ & 79.8 & 30.7 & 19.5 & 57.5 & 49.2 & 41.1 & 48.7 \\
    \arrayrulecolor{gray}\hline\arrayrulecolor{black}
    \rowcolor{gray!10} 15. & \textit{Ours (w/o \texttt{cPAE})} & B & \checkmark & 82.0 & 9.5 & 21.9 & 67.0 & 51.4 & 40.0 & 46.7 \\
    \rowcolor{gray!10} 16. & \textit{Ours (w/ \texttt{cPAE})} & B & \checkmark & 83.1 & 17.7 & 24.5 & 69.4 & 51.8 & \textbf{41.3} & \textbf{48.9} \\
    \bottomrule
\end{tabular}}
\vspace{-1.5mm}
\end{table}

\begin{table}[t]
\centering
\caption{Domain generalization evaluation. For SYNTHIA, mIoU computed over 16 categories. SO, ERM and LO indicate \textit{source-only}, empirical risk minimization and \textit{leave-one-out} respectively.
}
\label{tab:domaingeneralization}
\setlength{\tabcolsep}{8pt}
 \resizebox{1\columnwidth}{!}{
\begin{tabular}{lcccc}
    \toprule
         \multirow{2}{*}{Method} & \multicolumn{2}{c}{GTA5$\to$Cityscapes} & \multicolumn{2}{c}{SYNTHIA$\to$Cityscapes} \\
        \cmidrule(l{4pt}r{4pt}){2-3} \cmidrule(l{4pt}r{4pt}){4-5}
         & ResNet-101 & VGG16 & ResNet-101 & VGG16 \\
        \midrule
        IBN-Net \cite{pan2018IBN} & 37.1 & 34.7 & 35.6 & 33.0 \\
        ASG \cite{chen2020automated} & 38.8 & 35.4 & 36.9 & 34.2 \\
        DRPC \cite{yue2019domain} & 42.5 & - & 37.6 & 35.5 \\
        \rowcolor{gray!10} \textit{Ours (ERM) (SO)} & 43.1 & 38.9 & 40.1 & 36.9 \\
        \rowcolor{gray!10} \textit{Ours (LO++) (SO)} & \textbf{43.5} & \textbf{39.2} & \textbf{40.6} & \textbf{37.4} \\
    \bottomrule
\end{tabular}}
\vspace{-3mm}
\end{table}

\begin{table*}
    \centering
    \vspace{-6mm}
    \setlength{\tabcolsep}{4pt}
    \caption{Evaluating generalization and compatibility to online adaptation for GTA5$\to$Cityscapes models on Foggy-Cityscapes and NTHU-Cross-City datasets. 0.005, 0.01, and 0.02 indicate the levels of fog in the dataset and GT indicates ground truth segmentation maps. * indicates experiment reproduced by us using the released code of prior arts. We also show standard Cityscapes results for reference.}
     \resizebox{1\textwidth}{!}{
        \begin{tabular}{cllccc|cccc|ccccc}
        \toprule
         & \multirow{2}{*}{\#} & \multirow{2}{*}{Method} & \multirow{2}{*}{\parbox{3cm}{\centering \vspace{4pt} Access to \\ GTA5 $\vert$ Citysc.}} & \multicolumn{2}{c}{Cityscapes} & \multicolumn{4}{c}{Foggy-Cityscapes (19-class)} & \multicolumn{5}{c}{NTHU-Cross-City (13-class)} \\
        \cmidrule(l{4pt}r{4pt}){5-6} \cmidrule(l{4pt}r{4pt}){7-10} \cmidrule(l{4pt}r{4pt}){11-15}
         &  &  &  & 19-class & 13-class & 0.005 & 0.01 & 0.02 & \underline{Avg.} & Rio & Rome & Taipei & Tokyo & \underline{Avg.} \\
        \midrule
        \multirow{3}{*}{\parbox{3cm}{\centering Vendor-side \\ (GTA5)}} & 1. & BDL (w/o ST) \cite{li2019bidirectional} & \checkmark $\vert$ \checkmark (no GT) & 43.3 & 53.2 & 40.4 & 36.8 & 30.3 & \underline{35.8} & 38.9 & 42.2 & 42.2 & 41.2 & \underline{41.1} \\
          & 2. & FDA* (w/o ST) \cite{yang2020fda} & \checkmark $\vert$ \checkmark (no GT) & 42.7 & 51.9 & 42.1 & 40.3 & 35.3 & \underline{39.2} & 42.2 & 42.3 & 37.5 & 42.3 & \underline{41.0} \\
          & \cellcolor{gray!10}3. & \cellcolor{gray!10}\textit{Ours (vendor-side)} & \cellcolor{gray!10}\hspace{-11.7mm}\checkmark $\vert$ $\times$ & \cellcolor{gray!10}43.1 & \cellcolor{gray!10}51.5 & \cellcolor{gray!10}43.6 & \cellcolor{gray!10}42.4 & \cellcolor{gray!10}38.3 & \cellcolor{gray!10}\underline{\textbf{41.4}} & \cellcolor{gray!10}47.0 & \cellcolor{gray!10}48.7 & \cellcolor{gray!10}43.4 & \cellcolor{gray!10}44.5 & \cellcolor{gray!10}\underline{\textbf{45.9}} \\
        \midrule
        \multirow{5}{*}{\parbox{3cm}{\centering Client-side \\ ($\to$Citysc.)}} & 4. & ASN \cite{tsai2018learning} & \checkmark $\vert$ \checkmark (no GT) & 42.4 & 51.1 & 41.0 & 38.0 & 31.7 & \underline{36.9} & 41.8 & 44.5 & 37.5 & 41.9 & \underline{41.4} \\
          & 5. & MSL \cite{chen2019domain} & \checkmark $\vert$ \checkmark (no GT) & 46.4 & 54.5 & 44.3 & 40.9 & 34.2 & \underline{39.8} & 44.4 & 47.0 & 45.6 & 44.7 & \underline{45.4} \\
          & 6. & BDL \cite{li2019bidirectional} & \checkmark $\vert$ \checkmark (no GT) & 48.5 & 57.7 & 46.0 & 42.6 & 36.3 & \underline{41.6} & 44.1 & 47.1 & 47.5 & 44.3 & \underline{45.7} \\
          & 7. & FDA \cite{yang2020fda} & \checkmark $\vert$ \checkmark (no GT) & 48.8 & 57.8 & 47.6 & 45.2 & 39.1 & \underline{44.0} & 47.8 & 46.6 & 42.7 & 48.1 & \underline{46.3} \\
          & \cellcolor{gray!10}8. & \cellcolor{gray!10}\textit{Ours (client-side)} & \cellcolor{gray!10}$\times$ $\vert$ \checkmark (no GT) & \cellcolor{gray!10}53.4 & \cellcolor{gray!10}61.4 & \cellcolor{gray!10}51.7 & \cellcolor{gray!10}48.9 & \cellcolor{gray!10}42.3 & \cellcolor{gray!10}\underline{\textbf{47.6}} & \cellcolor{gray!10}47.1 & \cellcolor{gray!10}47.7 & \cellcolor{gray!10}45.7 & \cellcolor{gray!10}46.5 & \cellcolor{gray!10}\underline{\textbf{46.7}} \\
         \midrule
         \multirow{4}{*}{\parbox{3cm}{\centering Online Adapt. \\ ($\to$FoggyC / \\ $\to$NTHU)}} & 9. & CBST \cite{zou2018unsupervised} & $\times$ $\vert$ \checkmark (w/ GT) & -& - & - & - & - & - & 52.2 & 53.6 & 50.3 & 48.8 & \underline{51.2} \\
         & 10. & MSL \cite{chen2019domain} & $\times$ $\vert$ \checkmark (w/ GT) & -& - & - & - & - & - & 53.3 & 54.5 & 50.6 & 50.5 & \underline{52.2} \\
          & 11. &  CSCL \cite{dong2020cscl} & $\times$ $\vert$ \checkmark (w/ GT) & - & - & - & - & - & - & 53.8 & 54.8 & 51.4 & 51.0 & \underline{52.7} \\
         & \cellcolor{gray!10}12. & \cellcolor{gray!10}\textit{Ours (client-side)} & \cellcolor{gray!10}\hspace{-11.7mm}$\times$ $\vert$ $\times$ & \cellcolor{gray!10}- & \cellcolor{gray!10}- & \cellcolor{gray!10}53.6 & \cellcolor{gray!10}51.1 & \cellcolor{gray!10}45.9 & \cellcolor{gray!10}\underline{50.2} & \cellcolor{gray!10}54.3 & \cellcolor{gray!10}55.0 & \cellcolor{gray!10}51.6 & \cellcolor{gray!10}51.3 & \cellcolor{gray!10}\underline{\textbf{53.0}} \\
        \bottomrule
        \end{tabular}}
\label{tab:target_agnostic}
\end{table*}

\vspace{-5mm}
\subsubsection{Ablation study}
\vspace{-2mm}

Table \ref{table:ablation} reports a detailed ablation to independently analyse the components of the vendor and client side strategies. 

First, we evaluate the effectiveness of the proposed vendor-side strategies. For a fair comparison, we use a consistent client-side training for all the vendor-side ablations. As a baseline, we employ a standard (unaugmented) single-source-trained model. 
The ERM model gives an improvement of 3.2\% over the baseline. 
Next, we evaluate \textit{DE++} and observe an improvement of 0.4\%. 
\textit{LO++} gives a further improvement of 3.6\% over \textit{DE++}. This shows the clear superiority of \textit{LO++} over both ERM and \textit{DE++}.

Second, under client-side ablation, \texttt{cPAE} for pseudo-label extraction gives a boost of 1.8\%. Further, using \texttt{cPAE} for inference gives an additional 0.8-0.9\% improvement.

\vspace{-5mm}
\subsubsection{Analyzing cross-dataset generalization}
\vspace{-2mm}

Unlike prior arts which assume concurrent access to source and target (inculcates target-bias), our target-free vendor-side model is expected to generalize well to unseen targets. 
To this end, Table~\ref{tab:target_agnostic} shows our generalizability to other road-scene datasets, such as Foggy-Cityscapes \cite{SDHV18} and NTHU-Cross-City \cite{chen2017no}; before (\#1-3) and after (\#4-8) self-training on the related real domain, \ie Cityscapes. Among different variants, we achieve a superior average generalization even without concurrent access to samples from the related domain, Cityscapes. Note that concurrent access is beneficial to better characterize the domain gap.

\vspace{-5mm}
\subsubsection{Compatibility to online domain adaptation}
\vspace{-2mm}

Online adaptation~\cite{onlineda1,onlineda2} refers to a deployment setting where the model is required to continuously adapt to the current working conditions. The current state of the model may overcome its past domain-biases to perform the best at a given scenario. 
The proposed \textit{client-side} training can be seen as an online adaptation algorithm. Here, the frozen parameters of the multi-head \texttt{SoMAN} helps to retain task-specific knowledge while allowing adaptation to unlabeled samples from the new environment. In the last section of Table~\ref{tab:target_agnostic}, the initial Cityscapes adapted \texttt{SoMAN} is independently adapted to different secondary domains under Foggy-Cityscapes and NTHU-Cross-City. We also compare our results with recent Cityscapes$\to$NTHU-Cross-City works (\#9-12) that concurrently access labeled Cityscapes and unlabeled NTHU-Cross-City datasets.
The improved performance shows our compatibility to online adaptation.

\vspace{-3mm}
\section{Conclusion}
\vspace{-1mm}
We introduced a \textit{source-free} DA framework for semantic segmentation, recognizing practical scenarios where source and target data are not concurrently accessible. 
We cast the \textit{vendor-side} training as multi-source learning. Based on theoretical insights, we proposed \texttt{SoMAN} that balances generalization and specificity using the systematically selected \texttt{AG}s without access to the target.
To provide a strong support for the dense prediction task, \texttt{cPAE} is trained to denoise segmentation predictions and improve pseudo-label quality for \textit{client-side} source-free self-training.
Extending this framework to more DA scenarios involving category-shift can be a direction for future research.

\vspace{1mm}\noindent
\textbf{Acknowledgements.}
This work was supported by MeitY (Ministry of Electronics and Information Technology) project (No. 4(16)2019-ITEA), Govt. of India. 

\appendix

\vspace{2mm}
\noindent \textbf{\Large Supplementary Material}
\vspace{2mm}

In this supplementary, we provide detailed theoretical insights, extensive implementation details with additional qualitative and quantitative performance analysis. Towards reproducible research, we 
have released our code and trained network weights at our project page: \url{https://sites.google.com/view/sfdaseg}.

\noindent This supplementary is organized as follows:

\vspace{-2mm}

\begin{itemize}
\setlength{\itemindent}{-3mm}
\setlength{\itemsep}{-1mm}
    \item Section~\ref{sup:sec:notations}: Notations (Table~\ref{sup:tab:notations})
    \item Section~\ref{sup:sec:theoretical_insights}: Extended theoretical insights
    \vspace{-2mm}
    \begin{itemize}
        \setlength{\itemindent}{-3mm}
        \item Discussion on Result \textcolor{red}{1} (Sec.~\ref{sup:sec:discussion_result1})
        \item Augmentation selection criteria (Sec.~\ref{sup:sec:agselection_theory})
    \end{itemize}
    \item Section~\ref{sup:sec:implementation}: Implementation details
    \vspace{-2mm}
    \begin{itemize}
        \setlength{\itemindent}{-3mm}
        \item Experimental settings (Sec.~\ref{sup:sec:expt_setting}, Table~\ref{sup:tab:pae_arch})
        \item Vendor-side training (Sec.~\ref{sup:sec:vendorside}, Algo.~\ref{sup:algo:vendorside})
        \item Client-side training (Sec.~\ref{sup:sec:clientside}, Algo.~\ref{sup:algo:clientside})
        \item Optimization details (Sec.~\ref{sup:sec:optimization})
    \end{itemize}
    \vspace{-1mm}
    \item Section~\ref{sup:sec:analysis}: Analysis
    \vspace{-2mm}
    \begin{itemize}
        \setlength{\itemindent}{-3mm}
        \item Optimal choice of \texttt{AG}s (Sec.~\ref{sup:sec:agchoice}, Fig.~\ref{fig:DASS_sup_fig1}{\color{red}B}, Table~\ref{sup:tab:aug_candidates_priorarts})
        \item Empirical evaluation of Result \textcolor{red}{1} (Sec.~\ref{sup:sec:empirical_result1}, Table~\ref{sup:tab:empirical_result1})
        \item Impact of \texttt{cPAE} (Sec.~\ref{sup:sec:pae_impact}, Fig.~\ref{fig:DASS_sup_fig1}{\color{red}C})
        \item Time complexity analysis (Sec.~\ref{sup:sec:timecomplexity}, Table~\ref{sup:tab:trainingtime})
        \item Qualitative analysis (Sec.~\ref{sup:sec:qualitative}, Fig.~\ref{fig:DASS_sup_fig4}, \ref{fig:DASS_sup_fig5}, \ref{fig:DASS_sup_fig6}, \ref{fig:DASS_sup_fig3})
        \item Quantitative analysis (Sec.~\ref{sup:sec:quantitative}, Table~\ref{sup:tab:gta2city}, \ref{sup:tab:synthia2city})
    \end{itemize}
    \vspace{-1mm}
\end{itemize}{}

\section{Notations}
\label{sup:sec:notations}
\noindent We summarize the notations used in the paper in Table \ref{sup:tab:notations}. The notations are listed under 5 groups viz. Distributions, Datasets, Networks, Samples/Outputs and Theoretical.

\begin{table}[t]
    \centering
    \caption{\textbf{Notation Table}}
    \vspace{1mm}
    \setlength{\tabcolsep}{8pt}
    \resizebox{1\columnwidth}{!}{%
        \begin{tabular}{lcl}
        \toprule
            & Symbol & Description \\
        \midrule
        \multirow{4}{*}{\rotatebox[origin=c]{90}{Distribution}} & $p_s$ & Marginal source input distribution \\
         & $p_t$ & Marginal target input distribution \\
         & $p_y$ & Marginal output distribution \\
         & $p_{s_i}$ & Marginal $i^{\text{th}}$ aug. source input distr. \\
        \midrule
        \multirow{6}{*}{\rotatebox[origin=c]{90}{Datasets}} & $\mathcal{D}_s$ & Labeled source dataset \\
         & $\mathcal{D}_{s_i}$ & $i^{\text{th}}$-\texttt{AG} source dataset \\
         & $\mathcal{D}_{s_g}$ & \textit{Global} source dataset \\
         & $\mathcal{D}^-_{s_i}$ & $i^{\text{th}}$-\textit{leave-one-out} source dataset \\
         & $\mathcal{D}_t$ & Unlabeled target dataset \\
         & $\hat{\mathcal{D}}_t$ & Pseudo-labeled target dataset \\
        \midrule
        \multirow{5}{*}{\rotatebox[origin=c]{90}{Networks}} & $F$ & Shared CNN backbone \\
         & $H_g$ & Global output head \\
         & $H_i$ & $i^{\text{th}}$ non-global output head \\
         & $F_g$ & Backbone $F$ and first block of $H_g$ \\
         & $Q$ & Conditional Prior-enforcing AE \\
        \midrule
        \multirow{7}{*}{\rotatebox[origin=c]{90}{Samples / Outputs}} & $(x_s, y_s)$ & Labeled source sample \\
         & $\mathcal{T}_i(\cdot)$ & $i^\text{th}$ augmentation \\
         & $(x_{s_i}, y_s)$ & $i^\text{th}$ augmentation sample \\
         & $x_t$ & Unlabeled target sample \\
         & $(x_t, \hat{y}_t)$ & Pseudo-labeled target sample \\
         & $h_g$ & Output of $H_g \circ F$ \\
         & $h_i$ & Output of $H_i \circ F$ \\
        \midrule
        \multirow{4}{*}{\rotatebox[origin=c]{90}{Theoretical}} & $\epsilon_t(h)$ & Expected target error of hypothesis $h$ \\
         & $\alpha$ & Source convex combination weights \\
         & $\mathcal{H}^\alpha$ & Hypothesis subspace for particular $\alpha$ \\
         & $t_j$ & $j^\text{th}$ target domain \\
        \bottomrule
        \end{tabular}
        }
    \vspace{-5mm}
    \label{sup:tab:notations}
\end{table}

\section{Extended theoretical insights}
\label{sup:sec:theoretical_insights}

\subsection{Discussion on Result 1}
\label{sup:sec:discussion_result1}
To analyze possible solutions to the paradigm defined in Definition \textcolor{red}{1} of the paper, we introduce three configurations: \textbf{1)} ERM (empirical risk minimization) \ie weighting each multi-source domain equally, \textbf{2)} domain-experts++ (\textit{DE++}) \ie an ERM subspace together with $K$ subspaces formed from one specific domain each, \textbf{3)} leave-one-out++ (\textit{LO++}) \ie an ERM subspace together with $K$ subspaces formed by all domains except one each. We restate the result here:

\vspace{-6mm}
\begin{equation}
\label{sup:eqn:result1}
    \begin{split}
    \epsilon_t(h \in \mathcal{A}^\text{LO++}) \leq \epsilon_t(h \in \mathcal{H}^\text{ERM}) \\
    \epsilon_t(h \in \mathcal{A}^\text{DE++}) \leq \epsilon_t(h \in \mathcal{H}^\text{ERM})
    \end{split}
\end{equation}
\vspace{-4mm}

\noindent \textbf{Proof.} 
We give the proof by contradiction. Let the optimal hypothesis from \textit{LO++} have a higher target error than ERM \ie $\epsilon_t(h\in\mathcal{A}^\text{LO++})>\epsilon_t(h\in\mathcal{H}^\text{ERM})$. 
However, this implies that it cannot be the optimal hypothesis because an ERM hypothesis within \textit{LO++} has a lower target error. This is a contradiction which proves the result. The same can be shown for the \textit{DE++} case. In other words, the ERM subspace in \textit{DE++} or \textit{LO++} can provide the optimal hypothesis in the worst case and the equality in Eq. \ref{sup:eqn:result1} will hold.

\subsection{Augmentation selection criteria}
\label{sup:sec:agselection_theory}

From Eq. \textcolor{red}{3} of the paper, we know that the augmentation $\mathcal{T}_i$ modifies the non-causal domain-related factor $f_s$ by a weight $\gamma_i$ and introduces a new augmentation-related factor $f_i$ without altering the causal class factor $f_y$.
For a general augmentation, we cannot restrict $\gamma_i$ \ie $\gamma_i \in \mathbb{R}$. However, since we need to restrict the number of domains $K$, augmentations need to be filtered out. 
The important criteria for this filtering is the ability of the augmentation to simulate new domains while suppressing the original domain. Assuming that each augmentation is capable of introducing new domain-specific features, it is crucial for it to reduce the influence of the original domain. Thus, in Definition \textcolor{red}{2} in the paper, the selection criteria is $|\gamma_i| < 1$.

Due to the hypothetical nature of the decomposition of a sample into class-specific and domain-specific factors, it is not feasible to use the criteria on $\gamma_i$ directly.
However, models usually rely on both class-specific and domain-specific factors for prediction \cite{Peng2019DomainAL}. Thus, the performance of a standard single-source-trained model on augmented samples can be a good measure of the residual original domain dependency. 
An augmentation causing low performance for the pretrained model is likely to be suppressing the original domain factors.
In other words, the model is unable to latch onto the original domain factors for prediction. This gives the surrogate condition in Eq. \textcolor{red}{4} of the paper.
However, the candidate augmentations must be manually filtered to not select those that perturb the class-relevant factors since it is not ensured through this criteria.

\begin{figure*}
    \centering
    \includegraphics[width=\textwidth]{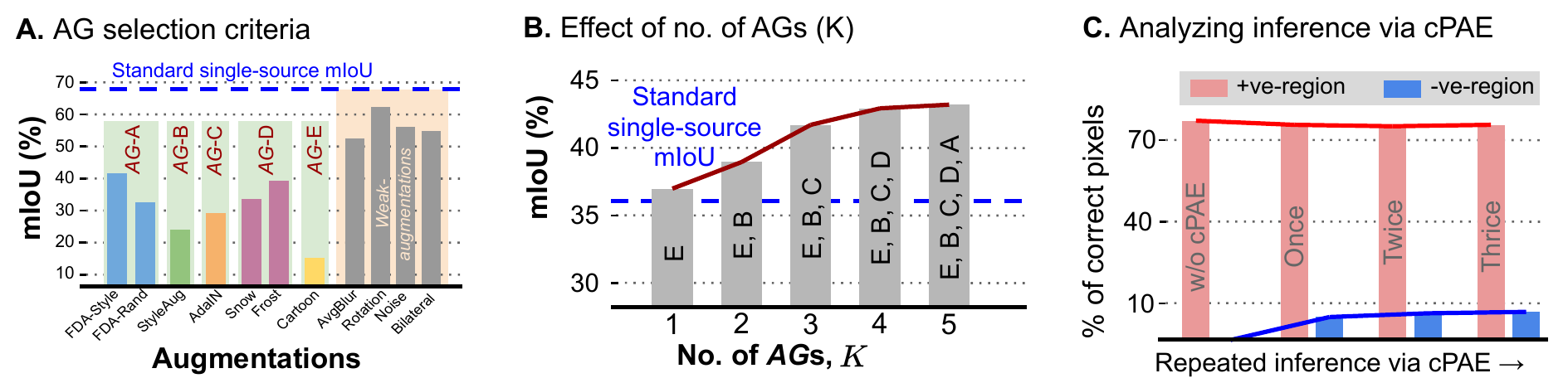}
    \caption{\textbf{A.} Selecting AGs from a set of candidate augmentations via inference through a standard single-source trained model (see Sec.~\ref{sup:sec:agchoice}). \textbf{B.} Performance of vendor-side trained models varying $K$ on Cityscapes. Performance saturates as $K$ reaches 5 (see Sec.~\ref{sup:sec:agchoice}). \textbf{C.} Impact of \texttt{cPAE} on correctly (+ve) and incorrectly (-ve) predicted regions on Cityscapes for a given model (see Sec.~\ref{sup:sec:pae_impact}).}
    \label{fig:DASS_sup_fig1}
\end{figure*}

\section{Implementation Details}
\label{sup:sec:implementation}
In this section, we describe the network architectures, datasets, the training process used for vendor-side and client-side training and other implementation details.

\subsection{Experimental settings}
\label{sup:sec:expt_setting}
\noindent \textbf{a) Network architectures.} Following \cite{li2019bidirectional, yang2020fda}, we employ 2 widely-used network architectures for the DA setting on semantic segmentation, DeepLabv2 \cite{chen2017deeplab} with ResNet101 \cite{he2016deep_resnet} backbone and FCN8s \cite{long2015fully} with VGG16 \cite{simonyan2014very} backbone. 
For DeepLabv2-ResNet101 and FCN8s-VGG16, we define the shared backbone $F$ upto \textit{Layer3} block and \textit{Conv4} block respectively. The remaining networks are taken as output head for both. Thus, at the inference stage, our model has exactly the same amount of parameters as used in the prior-arts. 
We use a fully convolutional architecture for the conditional Prior-enforcing Auto-Encoder (\texttt{cPAE}) with asymmetric encoder and decoder as shown in Table \ref{sup:tab:pae_arch}.

\vspace{1mm}
\noindent \textbf{b) Datasets.} We extensively evaluate the proposed approach on two popular synthetic-to-real benchmarks \textit{i.e.}, GTA5$\to$Cityscapes and SYNTHIA$\to$Cityscapes. 
For GTA5 \cite{richter2016playing}, we resize the image to $1280 \times 720$ before randomly cropping to $1024 \times 512$.  Whereas, for SYNTHIA \cite{ros2016synthia}, we resize to $1280 \times 760$ and random crop to $1024 \times 512$ following \cite{yang2020fda}. For Cityscapes \cite{cordts2016cityscapes}, we resize the image to $1024 \times 512$. For GTA5 we use 24500 images for training and 466 images for validation. Whereas, for SYNTHIA we use 9000 images for training and 400 images for validation. For client-side evaluation, we use Cityscapes training dataset for training and the standard validation set for testing 
\cite{yang2020fda}.
Following previous works~\cite{mei2020instance, Haoran_2020_ECCV, zou2019confidence}, we use multi-scale testing to report the final performance. 

\vspace{1mm}
\noindent \textbf{c) Augmentations.} We provide extra details about the \texttt{AG}s to enhance the reproducibility of our experiments.

\textbf{Aug-A}: We used images from a style transfer dataset \cite{huang2017adain} and release them with the code. For random noise, we sample uniformly from 0 to 255 for every pixel location.

\textbf{Aug-B}: We used this augmentation from the code release of \cite{jackson2019style} as provided. No controllable parameter available.

\textbf{Aug-C}: We set the strength of stylization ($\alpha$) to 0.3 which balances stylization and content preservation.

\textbf{Aug-D}: For snow and frost augmentation, we uniformly sample the severity between 1 and 3 (max. severity 5 possible in \cite{imgaug}) to balance stylization and content preservation.

\textbf{Aug-E}: No controllable parameter in cartoon \texttt{AG} \cite{imgaug}.

\subsection{Vendor-side training}
\label{sup:sec:vendorside}
The vendor-side training involves multi-head \texttt{SoMAN} training followed by \texttt{cPAE} training, as described in Algo.~\ref{sup:algo:vendorside}.

In \texttt{SoMAN} training, at each iteration, an image sampled from the source dataset $D_s$ is augmented using a random \texttt{AG}. Since we train each head in a \textit{leave-one-out} manner, the global head and $K-1$ non-global heads are trained at each iteration ({\color{red}L2-L14}). We update the parameters of each head using a separate optimizer ({\color{red}L13}). The momentum parameters in the optimizers adaptively scale the gradients thereby avoiding loss-scaling hyperparameters.

In \texttt{cPAE} training, we use the trained non-global heads from \texttt{SoMAN} to generate noisy seg-maps for denoising auto-encoder training. We augment source samples with a randomly chosen $i^{\text{th}}$ \texttt{AG} ({\color{red}L17-19}). Next, these are passed through the corresponding $i^{\text{th}}$ non-global head which was trained to be sensitive to that \texttt{AG}, thereby yielding noisy seg-maps ({\color{red}L20}). The \texttt{cPAE} predictions for these noisy seg-maps are used to compute the cross-entropy loss with the ground truth seg-maps. This loss is minimized using a SGD optimizer to update the \texttt{cPAE} parameters ({\color{red}L21-L23}).

\subsection{Client-side training}
\label{sup:sec:clientside}
The client-side training requires optimal head identification, pseudo-label extraction and self-training, as in Algo.~\ref{sup:algo:clientside}. 

In optimal head identification, for a given target, the head with the lowest $\epsilon_t(h)$ has to be chosen as per Result \textcolor{red}{1}. Since the computation of $\epsilon_t(h)$ is intractable, we choose a proxy \ie average self-entropy on the target training set (\textcolor{red}{L2}). Intuitively, the head closest to the target domain will be selected as it would be the most confident (lowest self-entropy).

In pseudo-label extraction, we first process the entire target training dataset ({\color{red}L3-L11}) and store the spatial class predictions ({\color{red}L8}) and the prediction probabilities ({\color{red}L9}). To avoid noisy predictions, we determine class-wise thresholds ({\color{red}L12-L16}) which are set at 33\% of the most confident predictions per class. Finally, we apply the class-wise thresholds ({\color{red}L17-23}) and assign an unlabeled `\textit{unknown}' class to the pixels which do not satisfy the threshold. These unlabeled pixels are not considered in the loss computation during training. 

In self-training, we use the pseudo-labeled target dataset to train in a supervised manner ({\color{red}L24-30}). Specifically, we train a block under the shared backbone $F$ ({\color{red}L29}) using the optimal head predictions and pseudo-labels. For DeepLabv2-ResNet101, this block is \textit{Layer3} while for FCN8s-VGG16, it is \textit{Conv3+Conv4} blocks. We find this to perform better than training the entire backbone $F$. We perform 3 rounds of offline pseudo-label extraction and self-training, following \cite{yang2020fda}. Further, we find that the performance does not increase with more than 3 rounds.

\subsection{Optimization details}
\label{sup:sec:optimization}
We implement our framework on PyTorch \cite{paszke2019pytorch}. Following \cite{li2019bidirectional}, for DeepLabv2-ResNet101, we use SGD optimizer with momentum 0.9, initial learning rate 2.5e-4 with a polynomial learning rate decay with power 0.9, and weight decay 5e-4. For FCN8s-VGG16, we use Adam \cite{kingma2014adam} optimizer with initial learning rate 1e-5 with step decay of 0.1 at every 20k steps. We train for 50k iterations each in vendor-side and in each round of self-training. With mixed precision, we use batch size 2 on a GTX1080Ti GPU.

\section{Analysis}
\label{sup:sec:analysis}
In this section, we analyze the choice of \texttt{AG}s, the empirical evaluation of Result \textcolor{red}{1} and the impact of \texttt{cPAE}.
We show extended quantitative and qualitative evaluations of our approach and training time comparisons with prior arts.

\subsection{Optimal choice of \texttt{AG}s}
\label{sup:sec:agchoice}
The choice of number of \texttt{AG}s and the \texttt{AG}s themselves is critical to the success of vendor-side training. 
We describe the \texttt{AG} candidates and those used by prior arts in Table~\ref{sup:tab:aug_candidates_priorarts}.
Firstly, it is important to choose diverse \texttt{AG}s to facilitate the learning of both domain-invariant and domain-specific features by the proposed \texttt{SoMAN}. 
Secondly, a higher number of \texttt{AG}s ($K$) and correspondingly non-global output heads incur additional computational cost in the vendor-side training. Thus, it is crucial to determine a low enough $K$ that yields a significant performance improvement. 

Towards the first, Fig.~\ref{fig:DASS_sup_fig1}{\color{red}A} shows the selection of diverse \texttt{AG}s from a set of candidates using the performance of a standard single-source trained model. This is according to Definition \textcolor{red}{2} in the paper using mIoU (task metric) with a threshold of 25\%. We observe that weaker augmentations that do not produce enough domain-shift get filtered out.

Towards the second, we analyze the effect of number of \texttt{AG}s on the performance in Fig.~\ref{fig:DASS_sup_fig1}{\color{red}B}. Particularly, we evaluate vendor-side trained models with varying number of \texttt{AG}s used during training. Since there are multiple ways to choose from the set of candidate \texttt{AG}s, we choose the most diverse ones first to ensure the best possible performance for a lower $K$. In other words, the \texttt{AG} giving the most deterioration in mIoU for a standard single-source trained model is chosen first. Fig.~\ref{fig:DASS_sup_fig1}{\color{red}A} shows that the order is E, B, C, D, A. Using this order to choose \texttt{AG}s for $K=\{1, 2, \dots, 5\}$, we observe that performance saturates as $K$ reaches $5$. Thus, we infer that the computational burden of adding more \texttt{AG}s would not result in a substantial performance improvement.

\begin{algorithm}[!t]
\caption{Pseudo-code for vendor-side training}
\label{sup:algo:vendorside}
\begin{algorithmic}[1]
\State \textbf{Input:} source dataset $\mathcal{D}_s$, standard single-source trained model $F_s, H_s$ \vspace{2mm}

\Statex \noindent $\triangleright$ \textit{Note that CE denotes class-weighted cross-entropy.} 
\vspace{2mm}

\Statex \underline{\textbf{Step 1: \textit{\texttt{SoMAN} training}}} \vspace{1mm}
\State Initialize $F, H_g, \{ H_i \}_{i=1}^K$ using $F_s, H_s$
\For{$iter < MaxIter$}:
    \State $x_s, y_s \leftarrow$ batch sampled from $\mathcal{D}_s$
    \State $i_1 \leftarrow \text{rand}(1, K)$
    \State $\tilde{x}_s \leftarrow$ $\mathcal{T}_{i_1}(x_s)$
    \State $h_g \leftarrow H_g(F(\tilde{x}_s))$
    \State Compute $\mathcal{L}_g \leftarrow$ CE($h_g, y_s$)
    \For{$i$ in range($K$) except $\{i_1\}$}:
        \State $h_i \leftarrow H_i(F(\tilde{x}_s))$
        \State Compute $\mathcal{L}_i \leftarrow$ CE($h_i, y_s$)
    \EndFor
    \State \textbf{update} $\theta_{H_g}, \{\theta_{H_i}\}_{i=1, i \neq i_1}^K$ by minimizing 
    \Statex \hspace{5mm} $\mathcal{L}_g, \{\mathcal{L}_i\}_{i=1, i \neq i_1}^K$ using separate optimizers
\EndFor \vspace{2mm}
\Statex \underline{\textbf{Step 2: \textit{\texttt{cPAE} training}}} \vspace{1mm}
\State Randomly initialize \texttt{cPAE} $Q$; freeze $F, H_g, \{ H_i \}_{i=1}^K$ from Step 1 training.
\For{$iter < MaxIter$}:
    \State $x_s, y_s \leftarrow$ batch sampled from $\mathcal{D}_s$
    \State $i \leftarrow \text{rand}(1, K)$ \Comment{\jnkc{Randomly choose an \texttt{AG}}}
    \State $x_{s_i} \leftarrow \mathcal{T}_i(x_s)$ \Comment{\jnkc{Augment $x_s$}}
    \State $h_i \leftarrow H_i(F(x_{s_i}))$\vspace{1mm}
    \State $\hat{h}_i \leftarrow Q(h_i, F_g(x_{s_i}))$
    \State Compute $\mathcal{L}_q \leftarrow$ CE($\hat{h}_i, y_s$)
    \State \textbf{update} $\theta_Q$ by minimizing $\mathcal{L}_q$ using SGD optimizer
\EndFor
\end{algorithmic}
\end{algorithm}

\begin{algorithm}
\caption{Pseudo-code for client-side training}
\label{sup:algo:clientside}
\begin{algorithmic}[1]

\State \textbf{Input:} Trained \texttt{SoMAN} ($F, H_g, \{ H_i \}_{i=1}^K$) and \texttt{cPAE} ($Q$) from vendor, unlabeled target dataset $\mathcal{D}_t$ \vspace{2mm}

\Statex \noindent $\triangleright$ \textit{Let $[\cdot]$ denote the indexing operation, $\cdot||\cdot$ denote the append operation, $|\cdot|$ denote the cardinality, and $C$ be the number of classes.} \vspace{3mm}

\Statex \underline{\textbf{Step 1: \textit{Optimal head identification}}} \vspace{1mm}
\State $i' \leftarrow \arg\min_{i \in \{g, [K] \}} \sum_{x \in \mathcal{D}_t}\{-\langle h_i, \log h_i \rangle \}$ where $h_i=H_i(F(x)) \, \forall \, i$ \Comment{\jnkc{Lowest average self-entropy}}

\Statex \underline{\textbf{Step 2: \textit{Pseudo-label extraction}}} \vspace{1mm}
\State $Y_p \leftarrow \{ \}$ \Comment{\jnkc{Empty ordered list}}
\State $W \leftarrow \{ \}$ \Comment{\jnkc{Empty ordered list}}
\State $X_p \leftarrow \{ \}$ \Comment{\jnkc{Empty ordered list}}
\For{$x$ in $\mathcal{D}_t$}:
    \State $\hat{h} = Q(H_{i'}(F(x)), F_g(x))$
    \State $y_p \leftarrow \arg \max_{c \in C} \hat{h}[c]$ \Comment{\jnkc{Class predictions}}
    \State $w \leftarrow \max_{c \in C} \hat{h}[c]$ \Comment{\jnkc{Predicted class probabilities}}
    \State $W, Y_p, X_p \leftarrow W \hspace{1mm} || \hspace{1mm} w, Y_p \hspace{1mm} || \hspace{1mm} y_p, X_p \hspace{1mm} || \hspace{1mm} x$
\EndFor

\State $t \leftarrow \{ \}$ \Comment{\jnkc{List of class-wise thresholds}}
\For{$c$ in range($C$)}:
    \State Store all prediction probabilities of class $c$ in $p_x$
    \Statex \hspace{4.4mm} $p_x \leftarrow W[Y_p == c]$
    \Statex \hspace{4.4mm} $p_x \leftarrow \text{sort}(p_x)$ 
    \State Set threshold at top 33\% most confident predictions
    \Statex \hspace{4.4mm} $t \leftarrow t \hspace{1mm} || \hspace{1mm} p_x[0.66 |p_x|]$
\EndFor

\State $\hat{\mathcal{D}}_t \leftarrow \{ \}$ \Comment{\jnkc{Empty ordered list}}
\For{$y_p, w, x_p$ in $Y_p, W, X_p$}:
    \For{$c$ in range($C$)}:
        \Statex \hspace{11mm}Assign class-id, $C+1$ representing `\textit{unknown}', 
        \Statex \hspace{11mm}to pixels with probability $<$ class threshold $t[c]$
        \State $y_p[(w < t[c])\&(y_p == c)] \leftarrow C+1$
    \EndFor
    \State $\hat{\mathcal{D}}_t \leftarrow \hat{\mathcal{D}}_t \hspace{1mm} || \hspace{1mm} (x_p, y_p)$
\EndFor \vspace{3mm}

\Statex \underline{\textbf{Step 3: \textit{Source-free self-training adaptation}}} \vspace{1mm}
\State Obtain $F, H_{i'}$ from vendor (or last self-training round)
\For{$iter < MaxIter$}:
    \State $x_t, y_t \leftarrow$ batch sampled from $\hat{\mathcal{D}}_t$
    \State $\hat{h} \leftarrow H_{i'}(F(x_t))$
    \State Compute $\mathcal{L}_t \leftarrow$ CE($\hat{h}, y_t$)
    \State \textbf{update} trainable parameters of $\theta_F$ by minimizing 
    \Statex \hspace{5mm}$\mathcal{L}_t$ using SGD optimizer
\EndFor \vspace{2mm}
\end{algorithmic}
\end{algorithm}

\begin{table}
    \centering
    \setlength{\tabcolsep}{1pt}
    \caption{\texttt{AG} candidates and augmentations used by prior arts. TF indicates target-free \ie whether augmentation requires target data.}
    \resizebox{1\columnwidth}{!}{
    \begin{tabular}{lcc}
        \toprule
        Method & TF & Description \\
        \midrule
        FDA \cite{yang2020fda} & $\times$ & Target images as reference for stylization in Fourier domain. \\
        LTIR \cite{kim2020learning} & $\times$ & Uses style-swap \cite{chen2016fast} with target images as reference. \\
        BDL \cite{li2019bidirectional} & $\times$ & Image-to-image translation for source$\to$target conversion. \\
        LDR \cite{yang2020label} & $\times$ & Image-to-image translation for target$\to$source conversion. \\
        \rowcolor{gray!10}\textit{Ours-A} & \checkmark & FDA \cite{yang2020fda} with random and style images as reference. \\
        \rowcolor{gray!10}\textit{Ours-B} & \checkmark & Stylization by randomly sampled style embedding \cite{jackson2019style}. \\
        \rowcolor{gray!10}\textit{Ours-C} & \checkmark & AdaIN \cite{huang2017adain} layers for stylization using reference images. \\
        \rowcolor{gray!10}\textit{Ours-D} & \checkmark & Varying levels of weather augmentations \cite{imgaug}: frost and snow. \\
        \rowcolor{gray!10}\textit{Ours-E} & \checkmark & Converts image into texture-less cartoon-like image \cite{imgaug}. \\
        \rowcolor{gray!10}\textit{Ours-W1} & \checkmark & Blurring using a $5\times 5$ average filter. \\
        \rowcolor{gray!10}\textit{Ours-W2} & \checkmark & Rotating image by an angle $\in [-15, 15]$ degrees. \\
        \rowcolor{gray!10}\textit{Ours-W3} & \checkmark & Adding random noise to the image. \\
        \rowcolor{gray!10}\textit{Ours-W4} & \checkmark & Edge-preserved smoothing using bilateral filtering. \\
        \bottomrule
    \end{tabular}}
    \label{sup:tab:aug_candidates_priorarts}
\end{table}

\subsection{Empirical evaluation of Result 1}
\label{sup:sec:empirical_result1}
To empirically evaluate Result \textcolor{red}{1} of the paper, we measure the performance of each \texttt{SoMAN} head for a variety of target scenarios as shown in Table~\ref{sup:tab:empirical_result1}. We observe that different heads of the \texttt{SoMAN} give the best performance in different target scenarios. Further, in most scenarios, at least one of the \textit{leave-one-out} heads performs better than ERM. This is in line with Result \textcolor{red}{1} \ie one of the leave-one-out heads gives a lower or equal target risk than ERM. 

For Foggy-Cityscapes (0.02) \cite{SDHV18}, we observe that ERM is optimal which can be considered as a worst-case scenario (large domain-shift). On the other hand, Foggy-Cityscapes (0.01) and (0.005) both have LO-E as the optimal head since they represent similar domain-shifts. Further, for NTHU-Cross-City \cite{chen2017no}, different heads are optimal for different cities since each city presents a different domain-shift.

\begin{table*}
    \centering
    \setlength{\tabcolsep}{15pt}
    \caption{ \textbf{Network architecture of cPAE.} \textbf{Conv*} denotes standard convolutional layer followed by a batch-normalization with Parametric-ReLU non-linearity, \textbf{Dconv} denotes standard convolutional layer with Dilation, \textbf{Tanh} denotes  hyperbolic tangent non-linearity, $\bigoplus$ denotes element-wise tensor addition, \textbf{Conv**} denotes standard convolutional layer followed by a batch-normalization with ReLU non-linearity, \textbf{Tconv*} denotes transpose convolutional layer followed by a batch-normalization with Parametric-ReLU non-linearity, {$||$} denotes channel-wise concatenation, \textbf{${F_g}$} consists of \texttt{SoMAN} backbone $F$ and first block of global head $H_g$ and input $x$ is an RGB image ($512\times 1024\times 3$).
    }
     \resizebox{1\textwidth}{!}{
     \begin{tabular}{rccccr}
    \toprule
    & Layer & Input & Type & Filter $\vert$ Stride $\vert$ Dilation & Output Size \\
    \midrule
    \multirow{12}{*}{\rotatebox[origin=c]{90}{Encoder}} & $C_1$ & $\hat{y}$ & Conv* & \hspace{4pt} $7\times7, 64$ $\vert$ 1 $\vert$ - \hspace{7pt} & $512\times1024\times64$ \\
    & $C_2$ & $C_1$ & Conv* & $3\times3, 128$ $\vert$ 2 $\vert$ - \hspace{5pt} & $256\times512\times128$\\
    & $C_3$ & $C_2$ & Conv* & $7\times7, 128$ $\vert$ 1 $\vert$ - \hspace{5pt} & $256\times512\times128$ \\
    & $C_4$ & $C_3$ & Conv* & $3\times3, 256$ $\vert$ 2 $\vert$ - \hspace{5pt} & $128\times256\times256$\\
    & $C_5$ & $C_4$ & Conv* & $7\times7, 256$ $\vert$ 1 $\vert$ - \hspace{5pt} & $128\times256\times256$ \\
    & $C_6$ & $C_5$ & Conv* & $3\times3, 512$ $\vert$ 2 $\vert$ - \hspace{5pt} & $64\times128\times512$\\
    & $C_7$ & $ C_6, F_g(x) $ & $||$ & - \hspace{5pt} & $64\times128\times2560$\\
    
    & $C_8$ & $C_7$ & Dconv & $3\times3, 512$ $\vert$ 1 $\vert$ 2 \hspace{3pt} & $64\times128\times512$\\
    & $C_9$ & $C_7$ & Dconv & $3\times3, 512$ $\vert$ 1 $\vert$ 4 \hspace{3pt} & $64\times128\times512$\\
    & $C_{10}$ & $C_7$ & Dconv & $3\times3, 512$ $\vert$ 1 $\vert$ 8 \hspace{3pt} & $64\times128\times512$\\
    & $C_{11}$ & $C_7$ & Dconv & $3\times3, 512$ $\vert$ 1 $\vert$ 16 & $64\times128\times512$\\
    & $C_{12}$ & $C_8, C_9, C_{10}, C_{11}$  & $\bigoplus$ & - & $64\times128\times512$\\
    & $C_{13}$ & $C_{12}$ & Conv+Tanh & $1\times1, 512$ $\vert$ 1 $\vert$ - \hspace{5pt} & $64\times128\times512$\\
    \midrule
    \multirow{7}{*}{\rotatebox[origin=c]{90}{Decoder}} & $C_{14}$ & $C_{13}$ & Conv** & $3\times3, 512$ $\vert$ 1 $\vert$ - \hspace{5pt} & $64\times128\times512$\\
    & $C_{15}$ & $C_{14}$ & Conv* & $3\times3, 512$ $\vert$ 1 $\vert$ - \hspace{5pt} & $64\times128\times512$\\
    & $C_{16}$ & $C_{15}$ & Conv* & $7\times7, 256$ $\vert$ 1 $\vert$ - \hspace{5pt}  & $64\times128\times256$\\
    & $C_{17}$ & $C_{16}$ & Tconv* & $3\times3, 256$ $\vert$ 2 $\vert$ - \hspace{5pt} & $128\times256\times256$\\
    & $C_{18}$ & $C_{17}$ & Conv* & $7\times7, 128$ $\vert$ 1 $\vert$ - \hspace{5pt} & $128\times256\times128$\\
    & $C_{19}$ & $C_{18}$ & Tconv* & \hspace{2pt} $3\times3, 64$ $\vert$ 2 $\vert$ - \hspace{5pt} & $256\times512\times64$\\
    & $C_{20}$ & $C_{19}$ & Conv & \hspace{2pt} $7\times7, 19$ $\vert$ 1 $\vert$ - \hspace{5pt} & $256\times512\times19$\\
    \midrule
    & Upsampling & $C_{20}$ & Interpolation & - & $512\times1024\times19$\\
    \bottomrule
    \end{tabular}
        }
\label{sup:tab:pae_arch}
\end{table*}

\subsection{Impact of \texttt{cPAE}}
\label{sup:sec:pae_impact}
We train the proposed \texttt{cPAE} as a denoising autoencoder to encourage spatial regularities in the segmentation predictions. In Fig.~\ref{fig:DASS_sup_fig1}{\color{red}C}, we analyze the effect of inference via \texttt{cPAE}. Particularly, we hypothesize that the \texttt{cPAE} should not distort regions that were correctly predicted by the segmentation network. This is desirable to retain the inductive bias in the absence of target labels. For a given segmentation network, we determine these regions, denoted as +ve-regions, and the regions where the segmentation network failed, denoted as -ve-regions. Next, we use the \texttt{cPAE} on the segmentation predictions and evaluate the performance in the two previously determined regions. We also evaluate repeated inference via \texttt{cPAE} \ie passing the output of the \texttt{cPAE} through the \texttt{cPAE} again. We observe that the performance in the +ve-region is almost the same while it improves in the -ve-region. Further, we observe that repeated inference via \texttt{cPAE} does not give any significant improvement. Thus, we resort to inferring via \texttt{cPAE} only once.

\subsection{Time complexity analysis}
\label{sup:sec:timecomplexity}
Our proposed client-side adaptation trains the shared backbone $F$ partially and does not require any additional networks like adversarial discriminators during the training. Further, we offer a simple adaptation pipeline without requiring access to source data. These factors lead to a lower training time (Table \ref{sup:tab:trainingtime}) for our client-side adaptation compared to prior arts while maintaining \textit{state-of-the-art} adaptation performance. This makes it suitable for practical and even online adaptation scenarios.

We perform the analysis (Table \ref{sup:tab:trainingtime}) by measuring the average time taken for forward pass, backward pass and the optimizer step (network weights update) for each method. For FDA and PCEDA, the time per iteration is higher since they train the entire model while using FFT-based augmentation and an image-to-image translation network respectively. 
For a fair comparison, we use batch size 1 (without automatic mixed precision) for all methods and evaluate on a machine with Intel Xeon E3-1200 CPU, 32GB RAM and a single 11GB NVIDIA GTX1080Ti GPU using Python 3, PyTorch 1.6 and CUDA 10.2.

\begin{table}
    \centering
    \setlength{\tabcolsep}{3pt}
    \caption{Empirical evaluation of Result \textcolor{red}{1} for vendor-side \texttt{SoMAN} heads with mIoU for various target scenarios. LO indicates leave-one-out head while ERM is the global head. 0.005, 0.01, and 0.02 indicate the levels of fog in the dataset. We observe that different heads are optimal for different target domains.}
    \resizebox{1\columnwidth}{!}{
    \begin{tabular}{lcccccccc}
        \toprule
        \multirow{2}{*}{Head}
        & \multirow{2}{*}{Cityscapes} & \multicolumn{3}{c}{Foggy-Cityscapes} & \multicolumn{4}{c}{NTHU-Cross-City} \\
        \cmidrule(l{4pt}r{4pt}){3-5} \cmidrule(l{4pt}r{4pt}){6-9}
         &  & 0.005 & 0.01 & 0.02 & Rio & Rome & Taipei & Tokyo \\
        \midrule
        ERM & 43.1 & 43.6 & 42.4 & \textbf{38.3} & 47.0 & 48.7 & 43.4 & 44.5 \\
        LO-A & 42.4 & 43.0 & {41.6} & 36.7 & 45.4 & \textbf{48.9} & 43.2 & 45.4 \\
        LO-B & 42.2 & 42.2 & 40.7 & 36.1 & \textbf{49.0} & 47.7 & 42.1 & {46.5} \\
        LO-C & 43.1 & 43.0 & 41.7 & 37.8 & 48.1 & 48.6 & 43.8 & \textbf{46.7} \\
        LO-D & \textbf{43.5} & 43.4 & 41.7 & 37.0 & 45.6 & 47.9 & \textbf{43.9} & 45.3 \\
        LO-E & 43.2 & \textbf{43.9} & \textbf{42.6} & 37.9 & 45.5 & {47.0} & 43.6 & 45.9 \\
        \bottomrule
    \end{tabular}}
    \label{sup:tab:empirical_result1}
\end{table}

\begin{table}
    \centering
    \setlength{\tabcolsep}{15pt}
    \caption{Training time comparison of adaptation step with prior arts. AN indicates whether additional networks are involved in the adaptation training.}
    \resizebox{1\columnwidth}{!}{
    \begin{tabular}{lcc}
        \toprule
        Method & AN & Training time (per iter.) (seconds) \\
        \midrule
        PCEDA \cite{yang2020phase} & \checkmark & 0.94 \\
        FDA \cite{yang2020fda} & $\times$ & 0.90 \\
        \rowcolor{gray!10}\textit{Ours} & $\times$ & \textbf{0.31} \\
        \bottomrule
    \end{tabular}}
    \label{sup:tab:trainingtime}
    \vspace{-4mm}
\end{table}

\begin{table*}
    \centering
    \vspace{3mm}
    \caption{\textbf{Quantitative evaluation on GTA5$\to$Cityscapes}. Performance on different segmentation architectures: A (DeepLabv2 ResNet-101), B (FCN8s VGG-16). SF indicates whether the method supports \textit{source-free} adaptation. \textit{Ours (V)} indicates use of our vendor-side \texttt{AG}s with prior art and * indicates reproduced by us using released code. We observe better or competitive performance on minority classes like motorcycle compared to non-source-free prior arts.}
    \setlength{\tabcolsep}{2pt}
    \resizebox{1\textwidth}{!}{%
    \begin{tabular}{lcccccccccccccccccccccc}
    \toprule
    Method & \rotatebox[origin=c]{60}{Arch.} & SF & \rotatebox[origin=c]{60}{road} & \rotatebox[origin=c]{60}{sidewalk} & \rotatebox[origin=c]{60}{building} & \rotatebox[origin=c]{60}{wall} & \rotatebox[origin=c]{60}{fence} & \rotatebox[origin=c]{60}{pole} & \rotatebox[origin=c]{60}{t-light} & \rotatebox[origin=c]{60}{t-sign} & \rotatebox[origin=c]{60}{vegetation} & \rotatebox[origin=c]{60}{terrain} & \rotatebox[origin=c]{60}{sky} & \rotatebox[origin=c]{60}{person} & \rotatebox[origin=c]{60}{rider} & \rotatebox[origin=c]{60}{car} & \rotatebox[origin=c]{60}{truck} & \rotatebox[origin=c]{60}{bus} & \rotatebox[origin=c]{60}{train} & \rotatebox[origin=c]{60}{motorcycle} & \rotatebox[origin=c]{60}{bicycle} & mIoU \\ 
    \midrule
    PLCA \cite{kang2020pixel} & A & $\times$ & 84.0 & 30.4 & 82.4 & 35.3 & 24.8 & 32.2 & 36.8 & 24.5 & 85.5 & 37.2 & 78.6 & 66.9 & 32.8 & 85.5 & 40.4 & 48.0 & 8.8 & 29.8 & 41.8 & 47.7 \\
    CrCDA \cite{huang2020contextualrelation} & A & $\times$ & 92.4 & 55.3 & 82.3 & 31.2 & 29.1 & 32.5 & 33.2 & 35.6 & 83.5 & 34.8 & 84.2 & 58.9 & 32.2 & 84.7 & 40.6 & 46.1 & 2.1 & 31.1 & 32.7 & 48.6 \\
    PIT \cite{lv2020cross} & A & $\times$ & 87.5 & 43.4 & 78.8 & 31.2 & 30.2 & 36.3 & 39.9 & \textbf{42.0} & 79.2 & 37.1 & 79.3 & {65.4} & 37.5 & 83.2 & 46.0 & 45.6 & 25.7 & 23.5 & 49.9 & 50.6 \\
    TPLD \cite{shin2020two-phase} & A & $\times$ & {94.2} & \textbf{60.5} & 82.8 & 36.6 & 16.6 & 39.3 & 29.0 & 25.5 & 85.6 & 44.9 & 84.4 & 60.6 & 27.4 & 84.1 & 37.0 & 47.0 & \textbf{31.2} & 36.1 & 50.3 & 51.2 \\
    RPT \cite{zhang2020transferring} & A & $\times$ & 89.7 & 44.8 & {86.4} & \textbf{44.2} & 30.6 & \textbf{41.4} & \textbf{51.7} & 33.0 & {87.8} & 39.4 & 86.3 & \textbf{65.6} & 24.5 & \textbf{89.0} & 36.2 & 46.8 & 17.6 & 39.1 & \textbf{58.3} & 53.2 \\
    FADA \cite{Haoran_2020_ECCV} & A & $\times$ & 91.0 & 50.6 & 86.0 & 43.4 & 29.8 & 36.8 & 43.4 & 25.0 & 86.8 & 38.3 & 87.4 & 64.0 & 38.0 & 85.2 & 31.6 & 46.1 & 6.5 & 25.4 & 37.1 & 50.1 \\
    IAST \cite{mei2020instance} & A & $\times$ & 94.1 & 58.8 & 85.4 & 39.7 & 29.2 & 25.1 & 43.1 & 34.2 & 84.8 & 34.6 & 88.7 & 62.7 & 30.3 & 87.6 & 42.3 & 50.3 & 24.7 & 35.2 & 40.2 & 52.2 \\
    \textit{Ours (V)} + FADA* & A & $\times$ & 91.2 & 51.0 & \textbf{86.6} & 43.6 & 30.3 & 37.1 & 43.7 & 25.2 & \textbf{87.9} & 40.2 & 88.2 & 64.7 & \textbf{38.4} & 85.5 & 32.0 & 46.8 & 6.6 & 25.9 & 37.5 & 50.6 \\
    \textit{Ours (V)} + IAST* & A & $\times$ & \textbf{94.8} & 59.4 & 86.2 & 40.5 & 29.5 & 25.5 & 43.8 & 34.7 & 85.9 & 34.9 & 89.5 & 63.4 & 30.8 & 88.3 & 42.6 & 50.7 & 25.3 & 35.7 & 40.9 & 52.8 \\
    \arrayrulecolor{gray}\hline\arrayrulecolor{black}
    URMA \cite{sivaprasad2021uncertainty} & A & \checkmark & 92.3 & 55.2 & 81.6 & 30.8 & 18.8 & 37.1 & 17.7 & 12.1 & 84.2 & 35.9 & 83.8 & 57.7 & 24.1 & 81.7 & 27.5 & 44.3 & 6.9 & 24.1 & 40.4 & 45.1 \\
    SRDA* \cite{bateson2020sourcerelaxed} & A & \checkmark & 90.5 & 47.1 & 82.8 & 32.8 & 28.0 & 29.9 & 35.9 & 34.8 & 83.3 & 39.7 & 76.1 & 57.3 & 23.6 & 79.5 & 30.7 & 40.2 & 0.0 & 26.6 & 30.9 & 45.8 \\
    \rowcolor{gray!10}\textit{Ours (w/o \texttt{cPAE})} & A & \checkmark & 90.9 & 48.6 & 85.5 & 35.3 & 31.7 & 36.9 & 34.7 & 34.8 & 86.2 & 47.8 & 88.5 & 61.7 & 32.6 & 85.9 & 46.9 & 50.4 & 0.0 & 38.9 & 52.4 & 51.6 \\
    \rowcolor{gray!10}\textit{Ours (w/ \texttt{cPAE})} & A & \checkmark & 91.7 & 53.4 & 86.1 & 37.6 & \textbf{32.1} & 37.4 & 38.2 & 35.6 & 86.7 & \textbf{48.5} & \textbf{89.9} & 62.6 & 34.3 & 87.2 & \textbf{51.0} & 50.8 & 4.2 & \textbf{42.7} & 53.9 & \textbf{53.4} \\
    \midrule
    BDL \cite{li2019bidirectional} & B & $\times$ & 89.2 & 40.9 & 81.2 & 29.1 & 19.2 & 14.2 & 29.0 & 19.6 & 83.7 & 35.9 & 80.7 & 54.7 & 23.3 & 82.7 & 25.8 & 28.0 & 2.3 & 25.7 & 19.9 & 41.3 \\
    LTIR \cite{kim2020learning} & B & $\times$ & 92.5 & 54.5 & \textbf{83.9} & \textbf{34.5} & \textbf{25.5} & \textbf{31.0} & 30.4 & 18.0 & 84.1 & 39.6 & 83.9 & 53.6 & 19.3 & 81.7 & 21.1 & 13.6 & \textbf{17.7} & 12.3 & 6.5 & 42.3 \\
    LDR \cite{yang2020label} & B & $\times$ & 90.1 & 41.2 & 82.2 & 30.3 & 21.3 & 18.3 & 33.5 & 23.0 & 84.1 & 37.5 & 81.4 & 54.2 & 24.3 & 83.0 & \textbf{27.6} & 32.0 & 8.1 & 29.7 & 26.9 & 43.6 \\
    FADA \cite{Haoran_2020_ECCV} & B & $\times$ & 92.3 & 51.1 & 83.7 & 33.1 & 29.1 & 28.5 & 28.0 & 21.0 & 82.6 & 32.6 & 85.3 & 55.2 & 28.8 & 83.5 & 24.4 & 37.4 & 0.0 & 21.1 & 15.2 & 43.8 \\
    PCEDA \cite{yang2020phase} & B & $\times$ & 90.2 & 44.7 & 82.0 & 28.4 & 28.4 & 24.4 & 33.7 & \textbf{35.6} & 83.7 & \textbf{40.5} & 75.1 & 54.4 & 28.2 & 80.3 & 23.8 & 39.4 & 0.0 & 22.8 & 30.8 & 44.6 \\
    \arrayrulecolor{gray}\hline\arrayrulecolor{black}
    SFDA \cite{liu2021source} & B & \checkmark & 81.8 & 35.4 & 82.3 & 21.6 & 20.2 & 25.3 & 17.8 & 4.7 & 80.7 & 24.6 & 80.4 & 50.5 & 9.2 & 78.4 & 26.3 & 19.8 & 11.1 & 6.7 & 4.3 & 35.8 \\
    \rowcolor{gray!10}\textit{Ours (w/o \texttt{cPAE})} & B & \checkmark & 90.1 & 44.2 & 81.7 & 31.6 & 19.2 & 27.5 & 29.6 & 26.4 & 81.3 & 34.7 & 82.6 & 52.5 & 24.9 & 83.2 & 25.3 & \textbf{41.9} & 8.6 & 15.7 & 32.2 & 43.4 \\
    \rowcolor{gray!10}\textit{Ours (w/ \texttt{cPAE})} & B & \checkmark & \textbf{92.9} & \textbf{56.9} & 82.5 & 20.4 & 6.0 & \textbf{30.8} & \textbf{34.7} & 33.2 & \textbf{84.6} & 17.0 & \textbf{88.9} & \textbf{62.3} & \textbf{30.7} & \textbf{85.1} & 15.3 & 40.6 & 10.2 & \textbf{30.1} & \textbf{50.4} & \textbf{45.9} \\
    \bottomrule
    \end{tabular}}
\label{sup:tab:gta2city}
\vspace{5mm}
\end{table*}

\begin{table*}
    \centering
    \caption{\textbf{Quantitative evaluation on SYNTHIA$\to$Cityscapes}. Performance on different segmentation architectures: A (DeepLabv2 ResNet-101), B (FCN8s VGG-16). mIoU and mIoU*
    are averaged over 16 and 13 categories respectively. SF indicates whether the method supports \textit{source-free} adaptation.}
    \setlength{\tabcolsep}{2pt}
    \resizebox{1\textwidth}{!}{%
    \begin{tabular}{lcccccccccccccccccccc}
    \toprule
    Method & \rotatebox[origin=c]{60}{Arch.} & SF & \rotatebox[origin=c]{60}{road} & \rotatebox[origin=c]{60}{sidewalk} & \rotatebox[origin=c]{60}{building} & \rotatebox[origin=c]{60}{wall*} & \rotatebox[origin=c]{60}{fence*} & \rotatebox[origin=c]{60}{pole*} & \rotatebox[origin=c]{60}{t-light} & \rotatebox[origin=c]{60}{t-sign} & \rotatebox[origin=c]{60}{vegetation} & \rotatebox[origin=c]{60}{sky} & \rotatebox[origin=c]{60}{person} & \rotatebox[origin=c]{60}{rider} & \rotatebox[origin=c]{60}{car} & \rotatebox[origin=c]{60}{bus} & \rotatebox[origin=c]{60}{motorcycle} & \rotatebox[origin=c]{60}{bicycle} & mIoU & mIoU* \\ 
    \midrule
    CAG \cite{zhang2019category} & A & $\times$ & 84.8 & 41.7 & \textbf{85.5} & - & - & - & 13.7 & 23.0 & \textbf{86.5} & 78.1 & \textbf{66.3} & 28.1 & 81.8 & 21.8 & 22.9 & 49.0 & - & 52.6 \\
    APODA \cite{yang2020adversarial} & A & $\times$ & 86.4 & 41.3 & 79.3 & - & - & - & 22.6 & 17.3 & 80.3 & 81.6 & 56.9 & 21.0 & 84.1 & 49.1 & 24.6 & 45.7 & - & 53.1 \\
    PyCDA \cite{lian2019constructing} & A & $\times$ & 75.5 & 30.9 & 83.3 & \textbf{20.8} & 0.7 & 32.7 & 27.3 & \textbf{33.5} & 84.7 & 85.0 & 64.1 & 25.4 & 85.0 & 45.2 & 21.2 & 32.0 & 46.7 & 53.3\\
    TPLD \cite{shin2020two-phase} & A & $\times$ & 80.9 & 44.3 & 82.2 & 19.9 & 0.3 & 40.6 & 20.5 & 30.1 & 77.2 & 80.9 & 60.6 & 25.5 & 84.8 & 41.1 & 24.7 & 43.7 & 47.3 & 53.5 \\
    USAMR \cite{zheng2019unsupervised} & A & $\times$ & 83.1 & 38.2 & 81.7 & 9.3 & 1.0 & 35.1 & 30.3 & 19.9 & 82.0 & 80.1 & 62.8 & 21.1 & 84.4 & 37.8 & 24.5 & 53.3 & 46.5 & 53.8 \\
    RPL \cite{zheng2020unsupervised} & A & $\times$ & 87.6 & 41.9 & 83.1 & 14.7 & 1.7 & 36.2 & 31.3 & 19.9 & 81.6 & 80.6 & 63.0 & 21.8 & 86.2 & 40.7 & 23.6 & 53.1 & 47.9 & 54.9 \\
    IAST \cite{mei2020instance} & A & $\times$ & 81.9 & 41.5 & 83.3 & 17.7 & \textbf{4.6} & 32.3 & 30.9 & 28.8 & 83.4 & 85.0 & 65.5 & 30.8 & 86.5 & 38.2 & 33.1 & 52.7 & 49.8 & 57.0 \\
    RPT \cite{zhang2020transferring} & A & $\times$ & 89.1 & 47.3 & 84.6 & 14.5 & 0.4 & \textbf{39.4} & \textbf{39.9} & 30.3 & 86.1 & 86.3 & 60.8 & 25.7 & \textbf{88.7} & 49.0 & 28.4 & \textbf{57.5} & 51.7 & 59.5 \\
    \arrayrulecolor{gray}\hline\arrayrulecolor{black}
    URMA \cite{sivaprasad2021uncertainty} & A & \checkmark & 59.3 & 24.6 & 77.0 & 14.0 & 1.8 & 31.5 & 18.3 & 32.0 & 83.1 & 80.4 & 46.3 & 17.8 & 76.7 & 17.0 & 18.5 & 34.6 & 39.6 & 45.0 \\
    \rowcolor{gray!10}\textit{Ours (w/o \texttt{cPAE})} & A & \checkmark & 89.0 & 44.6 & 80.1 & 7.8 & 0.7 & 34.4 & 22.0 & 22.9 & 82.0 & 86.5 & 65.4 & 33.2 & 84.8 & 45.8 & 38.4 & 31.7 & 48.1 & 55.5 \\
    \rowcolor{gray!10}\textit{Ours (w/ \texttt{cPAE})} & A & \checkmark & \textbf{90.5} & \textbf{50.0} & 81.6 & 13.3 & 2.8 & 34.7 & 25.7 & 33.1 & 83.8 & \textbf{89.2} & 66.0 & \textbf{34.9} & 85.3 & \textbf{53.4} & \textbf{46.1} & 46.6 & \textbf{52.0} & \textbf{60.1} \\
    \midrule
    PyCDA \cite{lian2019constructing} & B & $\times$ & 80.6 & 26.6 & 74.5 & 2.0 & 0.1 & 18.1 & 13.7 & 14.2 & 80.8 & 71.0 & 48.0 & 19.0 & 72.3 & 22.5 & 12.1 & 18.1 & 35.9 & 42.6 \\
    SD \cite{du2019ssf} & B & $\times$ & 87.1 & 36.5 & 79.7 & - & - & - & 13.5 & 7.8 & 81.2 & 76.7 & 50.1 & 12.7 & \textbf{78.0} & \textbf{35.0} & 4.6 & 1.6 & - & 43.4 \\
    FADA \cite{Haoran_2020_ECCV} & B & $\times$ & 80.4 & 35.9 & \textbf{80.9} & 2.5 & 0.3 & \textbf{30.4} & 7.9 & 22.3 & 81.8 & 83.6 & 48.9 & 16.8 & 77.7 & 31.1 & 13.5 & 17.9 & 39.5 & 46.0 \\
    BDL \cite{li2019bidirectional} & B & $\times$ & 72.0 & 30.3 & 74.5 & 0.1 & 0.3 & 24.6 & 10.2 & 25.2 & 80.5 & 80.0 & 54.7 & 23.2 & 72.7 & 24.0 & 7.5 & 44.9 & 39.0 & 46.1 \\
    PCEDA \cite{yang2020phase} & B & $\times$ & 79.7 & 35.2 & 78.7 & 1.4 & 0.6 & 23.1 & 10.0 & \textbf{28.9} & 79.6 & 81.2 & 51.2 & \textbf{25.1} & 72.2 & 24.1 & \textbf{16.7} & \textbf{50.4} & 41.1 & 48.7 \\
    \arrayrulecolor{gray}\hline\arrayrulecolor{black}
    \rowcolor{gray!10}\textit{Ours (w/o \texttt{cPAE})} & B & \checkmark & 88.5 & 45.4 & 79.8 & 2.8 & 2.2 & 27.4 & 18.4 & 25.4 & 82.4 & 83.6 & 55.9 & 12.1 & 72.8 & 25.6 & 3.5 & 12.9 & 40.0 & 46.7 \\
    \rowcolor{gray!10}\textit{Ours (w/ \texttt{cPAE})} & B & \checkmark & \textbf{89.9} & \textbf{48.8} & \textbf{80.9} & \textbf{2.9} & \textbf{2.5} & 28.1 & \textbf{19.5} & 26.2 & \textbf{83.7} & \textbf{84.9} & \textbf{57.4} & 17.8 & 75.6 & 28.9 & 4.3 & 17.2 & \textbf{41.3} & \textbf{48.9} \\
    \bottomrule
    \end{tabular}}
\label{sup:tab:synthia2city}
\vspace{4mm}
\end{table*}

\begin{figure*}
    \centering
    \includegraphics[width=\textwidth]{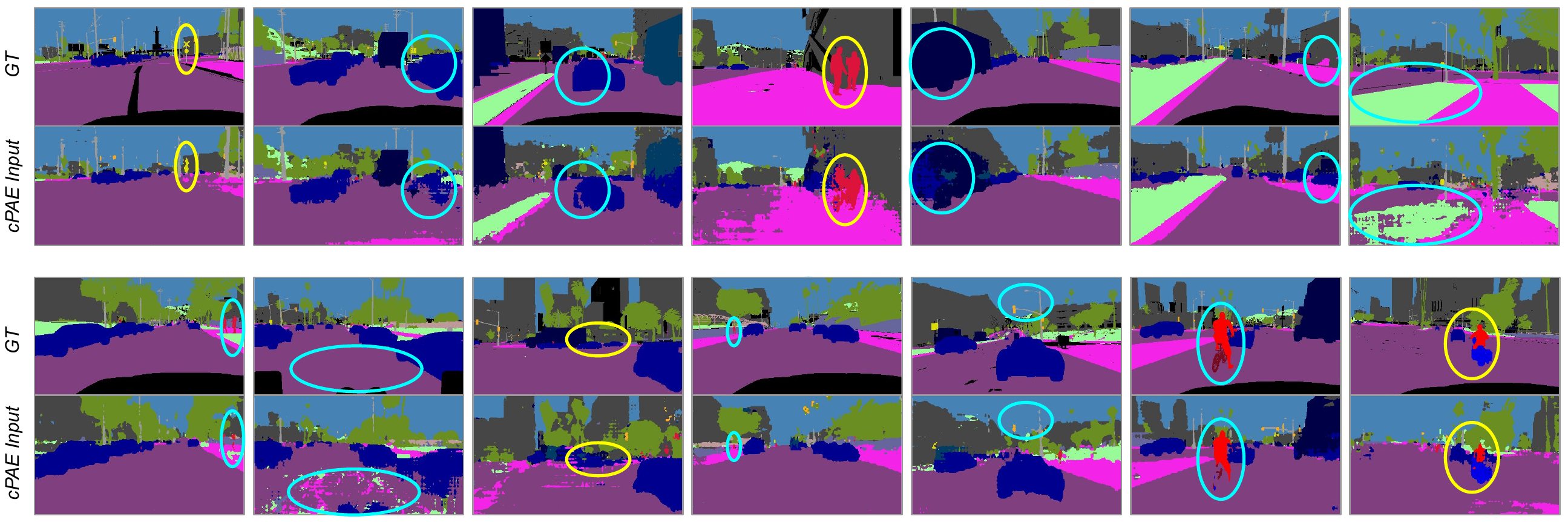}
    \caption{Paired samples for \texttt{cPAE} training. \texttt{cPAE} is trained as a denoising autoencoder to encourage structural regularity in segmentation predictions and alleviate merged-region (yellow circle) and splitted-region (blue circle) problems. \textit{Best viewed in color.}
    }
    \label{fig:DASS_sup_fig4}
\end{figure*}

\begin{figure*}
    \centering
    \includegraphics[width=\textwidth]{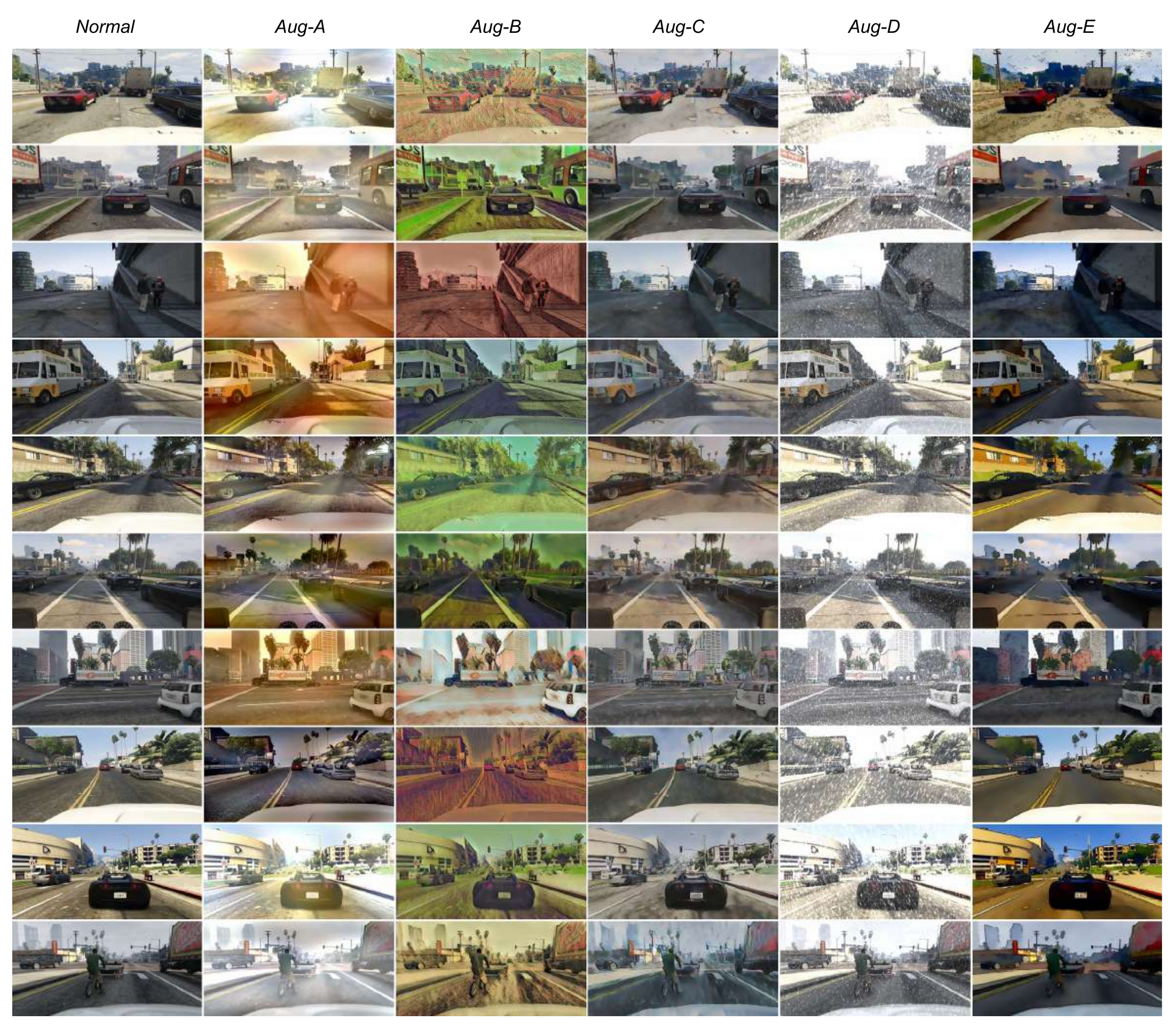}
    \caption{Examples of \texttt{AG}s applied to GTA5 dataset images. Notice the diversity in the augmentations. \textit{Best viewed in color.}}
    \vspace{-4mm}
    \label{fig:DASS_sup_fig5}
\end{figure*}

\begin{figure*}
    \centering
     \vspace{7mm}
    \includegraphics[width=\textwidth]{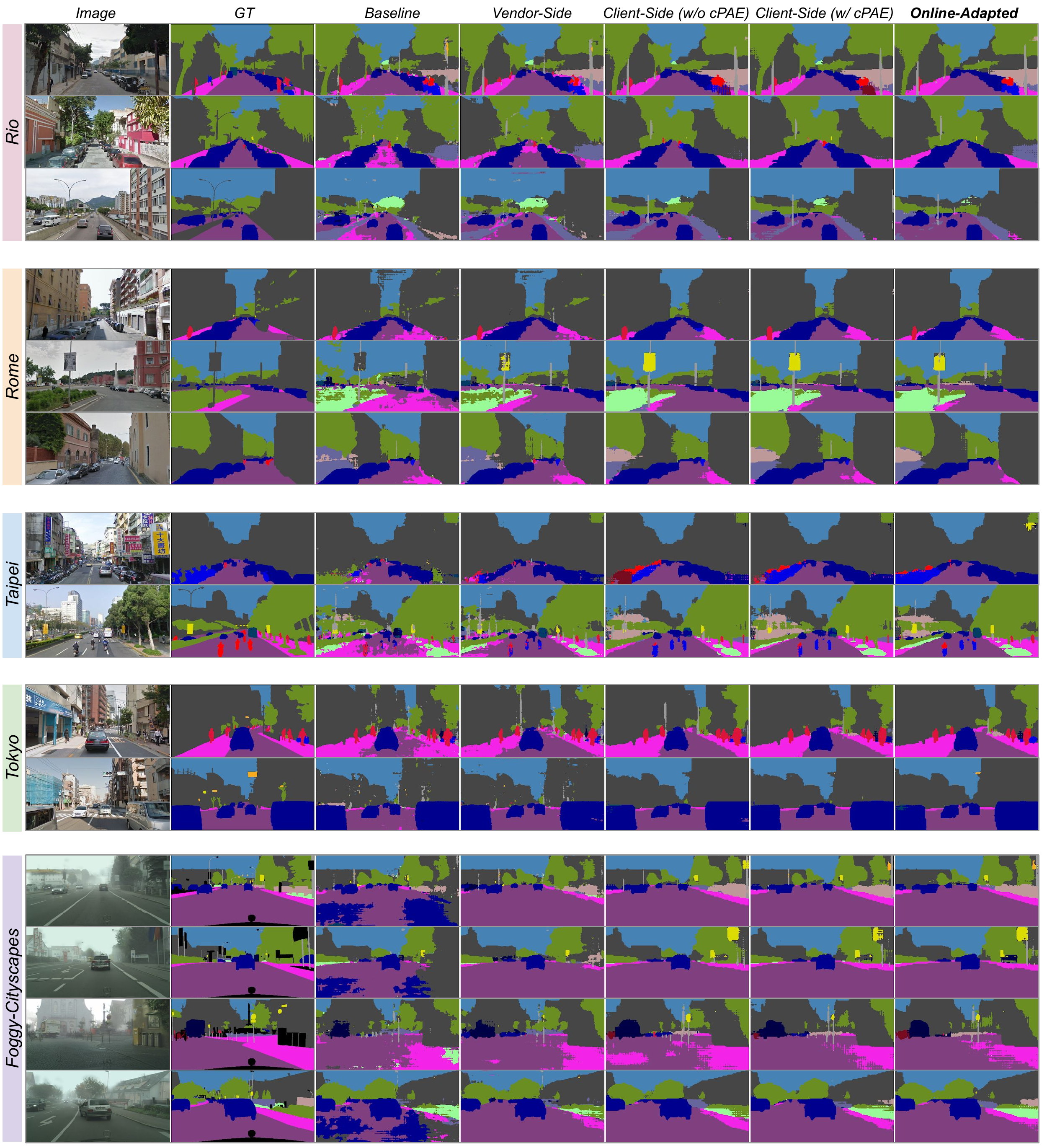}
    \caption{Qualitative evaluation of GTA5$\to$Cityscapes and online adapted models on NTHU-Cross-City and Foggy-Cityscapes datasets. The performance generally improves from vendor-side to client-side to online-adapted model. \textit{Best viewed in color.}
    }
    \vspace{7mm}
    \label{fig:DASS_sup_fig6}
\end{figure*}

\begin{figure*}
    \centering
    \includegraphics[width=\textwidth]{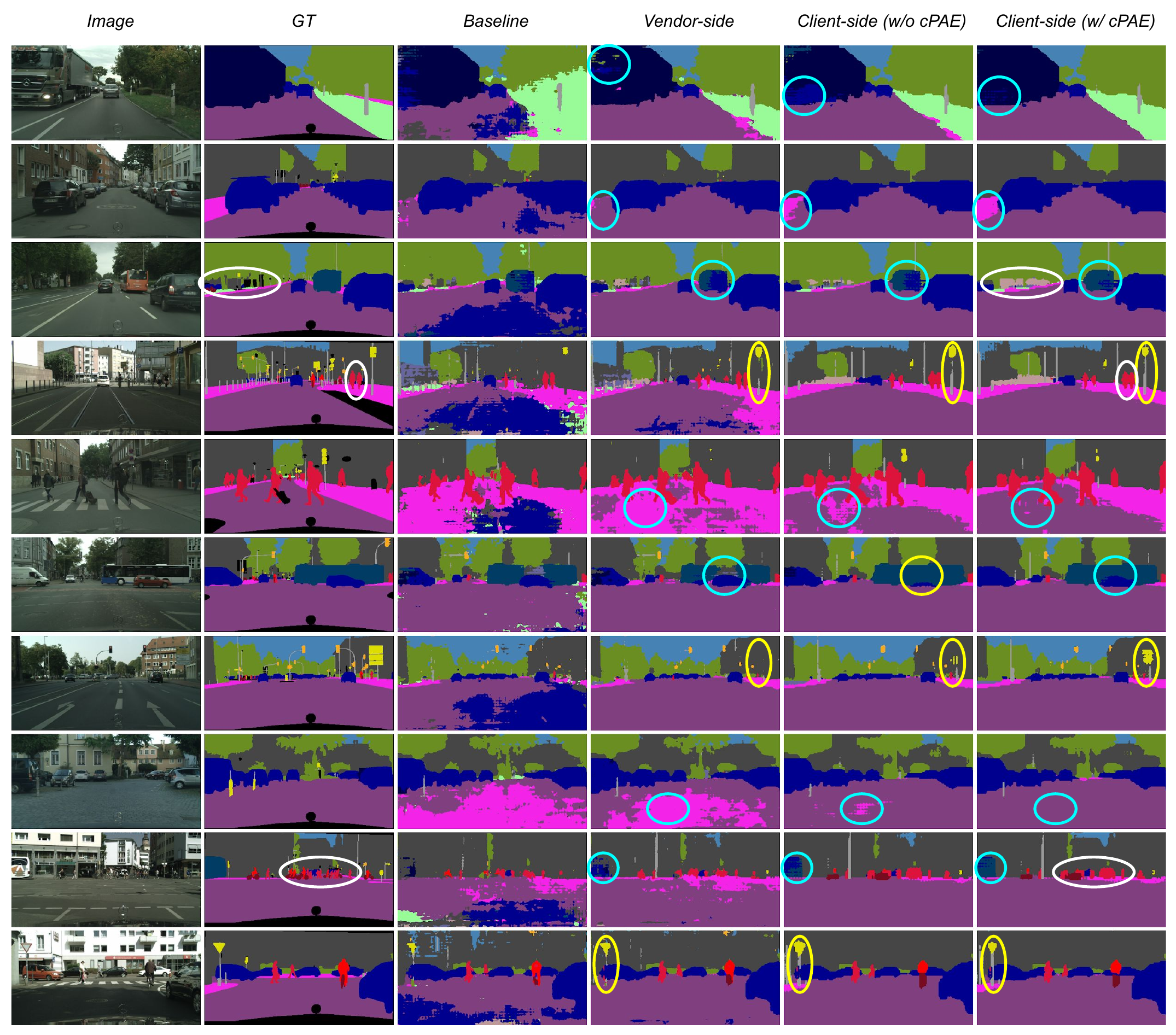}
    \caption{Qualitative evaluation of the proposed approach. Vendor-side model generalizes better than baseline but performs worse than client-side due to the domain gap. Inculcating prior knowledge from \texttt{cPAE} structurally regularizes the predictions and overcomes merged-region (yellow circle) and splitted-region (blue circle) problems. Some failure cases are also shown (white circle). \textit{Best viewed in color.}}
    \label{fig:DASS_sup_fig3}
\end{figure*}

\subsection{Qualitative analysis}
\label{sup:sec:qualitative}
We provide extended qualitative results of our proposed approach on GTA5$\to$Cityscapes \cite{richter2016playing, cordts2016cityscapes} in Fig.~\ref{fig:DASS_sup_fig3}. Further, we show examples of the paired samples used for the training of \texttt{cPAE} in Fig.~\ref{fig:DASS_sup_fig4} and examples of the devised \texttt{AG}s applied to the GTA5 dataset in Fig.~\ref{fig:DASS_sup_fig5}. We also show qualitative results of online adaptation to NTHU-Cross-City \cite{chen2017no} and Foggy-Cityscapes \cite{SDHV18} in Fig.~\ref{fig:DASS_sup_fig6}.

We also observe some failure cases in Fig.~\ref{fig:DASS_sup_fig3} (indicated by white circles) where merged-region problems occur for smaller-sized classes in the scene. More explicit ways of inculcating shape priors may improve the performance further. We plan to explore this direction in our future work.

\subsection{Quantitative analysis}
\label{sup:sec:quantitative}
We provide extended quantitative results on the GTA5$\to$Cityscapes and SYNTHIA$\to$Cityscapes \cite{ros2016synthia} benchmarks for semantic segmentation in Tables~\ref{sup:tab:gta2city}, \ref{sup:tab:synthia2city}. We obtain \textit{state-of-the-art} performance across all settings even against non-source-free approaches.

\clearpage

{
\small
\bibliographystyle{ieee_fullname}
\bibliography{egbib}
}

\end{document}